\documentclass[10pt,journal,compsoc]{IEEEtran}

\usepackage{color}
\usepackage{epsfig}
\usepackage{graphicx}
\usepackage{amsmath}
\usepackage{amssymb}
\usepackage{multirow}
\usepackage{epstopdf}
\usepackage{subfigure}
\usepackage{ragged2e}
\usepackage{url}
\usepackage{booktabs}
\usepackage{enumitem}
\usepackage{algorithm}
\usepackage{algpseudocode}
\usepackage{makecell}
\usepackage{hyperref}
\usepackage{xcolor}
\usepackage{pdfpages}

\hypersetup{
colorlinks=true,
linkcolor=black
}

\newcommand{\etal}{\textit{et al.}}
\newcommand{\etc}{\textit{etc}}

\newcommand{\ys}{\textcolor{black}}
\newcommand{\lyf}{\textcolor{black}}

\ifCLASSOPTIONcompsoc
  \usepackage[nocompress]{cite}
\else
  \usepackage{cite}
\fi

\begin{document}

\title{Language-based Image Colorization: A Benchmark and Beyond}

\author{Yifan~Li, 
        Shuai~Yang,~\IEEEmembership{Member,~IEEE},
        and~Jiaying~Liu,~\IEEEmembership{Fellow,~IEEE}
}


\IEEEtitleabstractindextext{
\begin{abstract}
\justifying
Image colorization aims to bring colors back to grayscale images. Automatic image colorization methods, which requires no additional guidance, struggle to generate high-quality images due to color ambiguity, and provides limited user controllability. Thanks to the emergency of cross-modality datasets and models, language-based colorization methods are proposed to fully utilize the efficiency and flexibly of text descriptions to guide colorization. In view of the lack of a comprehensive review of language-based colorization literature, we conduct a thorough analysis and benchmarking. We first briefly summarize existing automatic colorization methods. Then, we focus on language-based methods and point out their core challenge on cross-modal alignment. We further divide these methods into two categories: one attempts to train a cross-modality network from scratch, while the other utilizes the pre-trained cross-modality model to establish the textual-visual correspondence. Based on the analyzed limitations of existing language-based methods, we propose a simple yet effective method based on distilled diffusion model. Extensive experiments demonstrate that our simple baseline can produces better results than previous complex methods with 14 times speed up. To the best of our knowledge, this is the first comprehensive review and benchmark on language-based image colorization field, providing meaningful insights for the community. The code is available at \href{https://github.com/lyf1212/Color-Turbo}{https://github.com/lyf1212/Color-Turbo}.
\end{abstract}

\begin{IEEEkeywords}
    Image colorization, deep neural network, cross-modality alignment, language-based generation
\end{IEEEkeywords}}

\maketitle

\IEEEdisplaynontitleabstractindextext
\IEEEpeerreviewmaketitle

\ifCLASSOPTIONcompsoc
\IEEEraisesectionheading{\section{Introduction}\label{sec:introduction}}
\else
\section{Introduction}
\fi

\IEEEPARstart{I}{mage} colorization aims at colorizing a grayscale image with plausible colors, which has a wide range of applications on historical photos restoration and the film industry. 
However, image colorization is an ill-posed problem as it involves inferring a three-channel color image from a single-channel luminance image, which may have multiple reasonable solutions. It is still a challenge to find the optimal high-quality solution. 

\begin{figure}[t]
    \centering
    \includegraphics[width=0.9\linewidth]{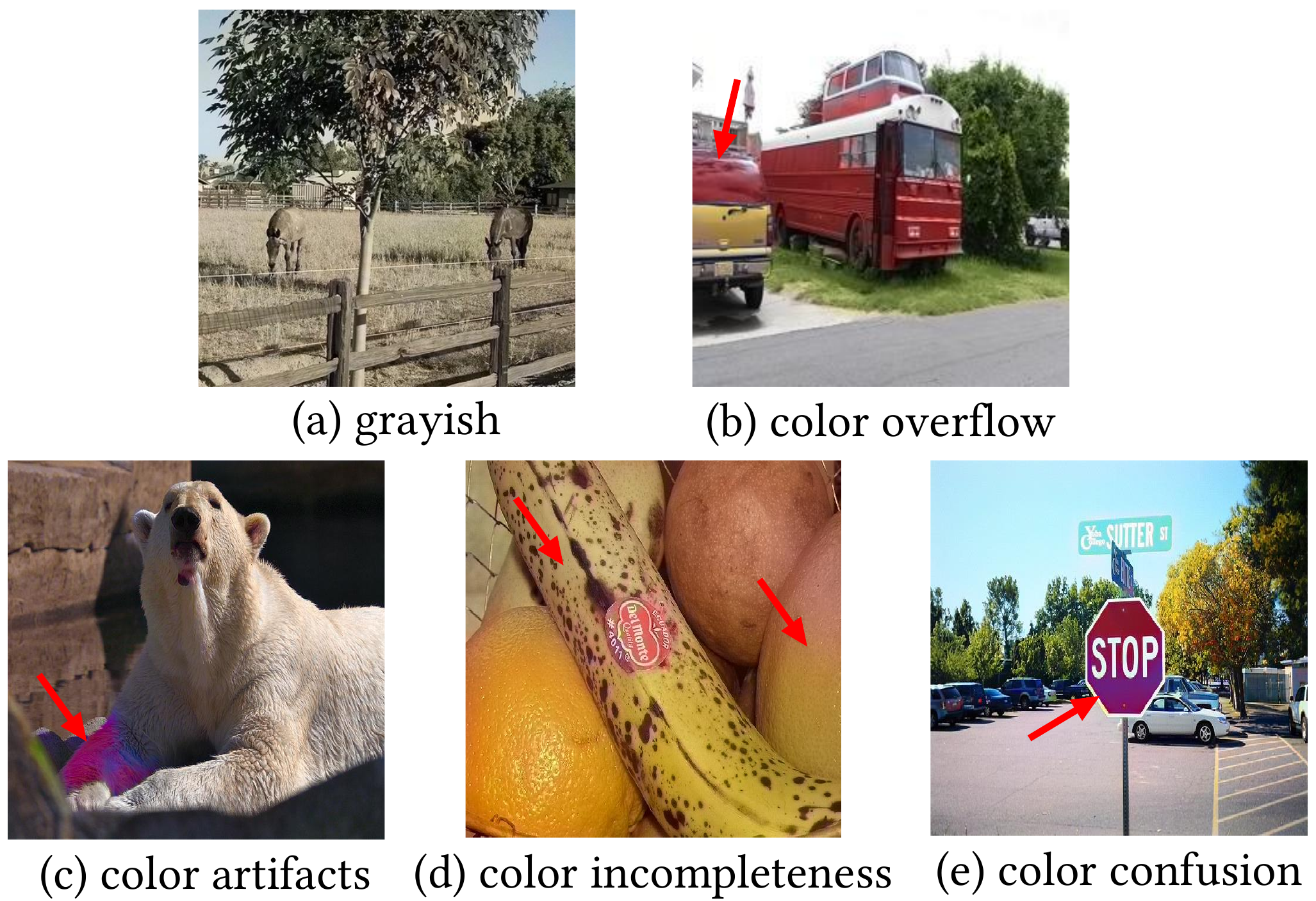}\vspace{-3mm}
    \caption{Five typical color distortions of automatic colorization methods. From (a) to (e), the image is generated by HistoryNet~\cite{HistoryNet}, CT2~\cite{ct2}, BigColor~\cite{bigcolor}, CIC~\cite{cic} and DDColor~\cite{ddcolor}. 
    }
    \label{discuss_uncond}
    \vspace{-3mm}
\end{figure}

Image colorization can be divided into two categories, the unconditional methods and the conditional methods. Unconditional colorization methods, or automatic colorization methods, attempt to colorize grayscale images without any additional guidance. In the deep learning era, it is completely realized by learning a mapping from hundreds of thousands of gray-color images pairs to fit the target image distribution with the strong learning ability of deep models~\cite{cic, deep_color, ct2, colorformer}. 
Although advanced architectures such as Generative Adversarial Network (GAN~\cite{gan}) and diffusion models have been exploited, due to the high ill-posed nature of the task, these models may still suffer from color ambiguity and limited controllability, leading to color overflows, incompleteness and artifacts. 
Moreover, since a gray pixel value corresponds to a variety of possible color choices, the models tend to average the possible solutions and yield degraded under-saturated grayish results.
Besides, such color ambiguity could cause color confusion, \textit{i.e.}, two reasonable colors occur on a single object. We visualize the above color distortions in Fig.~\ref{discuss_uncond}.

As for conditional colorization, conditional methods take diverse kinds of instructions, such as text descriptions~\cite{lcode, lcoder, lcoins, lcad, coco-lc}, hint points~\cite{iColoriT}, color strokes~\cite{unicolor, CtrlColor}, palette~\cite{palgan}, \etc., to guide the colorization process with more deterministic cues, therefore alleviating color ambiguity. 

\begin{figure}[t]
    \centering
    \includegraphics[width=1\linewidth]{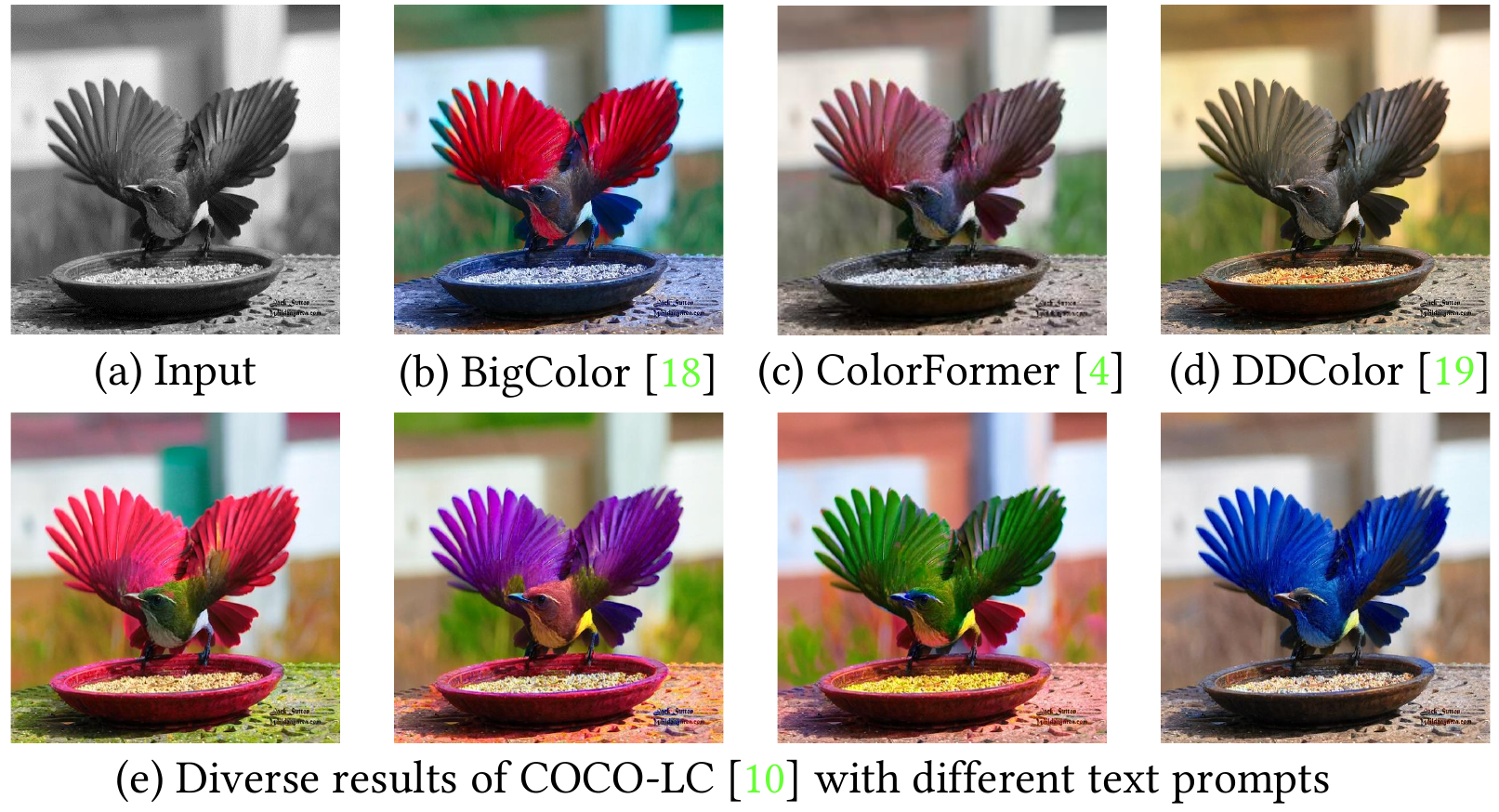}\vspace{-3mm}
    \caption{Comparison of automatic and language-based methods. Automatic colorization methods~\cite{bigcolor,colorformer,ddcolor} tend to produce overflowed, under-saturated and incomplete results. Language-based method~\cite{coco-lc} can not only generate high quality results with deterministic color cues, but also produce diverse results according to different user commands.}
    \label{compare_uncond_cond}
    \vspace{-5mm}
\end{figure}

Compared with text descriptions, hint points, color strokes and palette have higher learning and usage costs for users: users need to determine where to draw and summarize what color values to assign.
With the development of cross-modality datasets and models~\cite{clip, schuhmann2022laion}, language-based colorization begins to attract more attention for language is the most efficient and natural way for human to communicate and deliver information. 
This kind of methods shows its potential to achieve a promising balance between controllability and usability on conditional colorization problem, as shown in Fig.~\ref{compare_uncond_cond}, where automatic methods produce under-saturated results, and language-based methods can generate high quality results according to user commands.

\ys{However, there is still a lack of a thorough literature review and analysis on this relatively new and active research problem.} 
Therefore, this paper mainly focuses on language-based colorization and conduct a comprehensive survey on both methodology and performance evaluation.
Besides, thanks to the emergency of image captioning~\cite{blip}, language-based colorization methods can be applied to automatic colorization in a zero-shot manner by automatically annotating the gray image with text. Under such unified perspective, to build a comprehensive benchmark, we also conduct a brief review on existing automatic colorization methods, and benchmark on both automatic and language-based colorization tasks. 
\ys{We also notice that some one-for-all methods~\cite{gdp, ddrm, dps, vps} use pre-trained diffusion models as general image priors for various restoration tasks, including colorization. Since these one-for-all methods are not specially designed for image colorization, this paper does not discuss and benchmark them.}


After thorough review on previous methods, we argue that the main challenge of language-based colorization is to establish an accurate correspondence between text descriptions and grayscale images, which requires strong cross-modality capabilities.
A part of previous language-based methods~\cite{lcode, lcoder, lcoins} make efforts on conducting such alignment with the help of a manual annotated dataset, handcrafted cross-modality mechanism \ys{and pre-trained large language models (LLMs)~\cite{bert}}. 
These methods first find correspondence between object words and their proper locations in the grayscale images, and then assign color words to certain images regions based on the relationship of them and object words in the text descriptions.
\ys{Later, pre-trained cross-modality models like CLIP~\cite{clip} is explored to explicitly align color words and grayscale images~\cite{unicolor}.
However, it is still challenging to realize spatially fine-grained colorization ability on open-world color words.}


Recently, fueled by the success in large-scale text-to-image generation models~\cite{sd, sd-turbo}, researchers~\cite{CtrlColor, coco-lc, lcad} leverage the strong generative prior of Stable Diffusion~\cite{sd} and formulate colorization as a conditional generation task. \ys{As this text-to-image model inherently has powerful cross-modality performance,}
the main challenge now lies in how to achieve the \lyf{structural} consistency between grayscale images and the generated color images while maintaining the generative ability. 
\ys{We summarize four types of representative condition insertion paradigms adopted by existing methods for structural consistency, revealing that 
the four paradigms have diffferent injection intensities and granularity, and it is still hard to maintain a balance between color generation and structural consistency.}

Based on our literature review, we notice that current language-based colorization methods are still limited by long inference time and unstable colorization performance including \lyf{under-/over-saturation}, color overflows and color artifacts. We analyze the causes of these issues and correspondingly propose a simple yet effective language-based image colorization method based on the distilled diffusion models~\cite{sd-turbo}, called \textbf{Color-Turbo}, hoping to build a strong and efficient baseline for the community. Extensive experiments verify the promising performance of proposed method compared to state-of-the-art methods.

In this paper, we conduct a benchmark of existing language-based colorization methods. 
We begin by contributing a comprehensive survey of both automatic and language-based image colorization methods.
Then, we present an analysis in depth on different condition insertion strategies and a discussion on future possible directions. To the best of our knowledge, we are the first to provide a thorough survey that comprehensively reviews and evaluates language-based image colorization literature. We believe our extensive work can benefit the research community and bring rich insights for future work.

Furthermore, in our benchmark, we experimentally find that common metrics cannot reflect human perference accurately.
To provide a comprehensive evaluation protocol to measure previous image colorization methods, we propose a new metric based on original FID~\cite{fid} metric, 
named hue-invariant FID,
which more accurately measures the colorization quality without being affected by the hue deviation. 
We show in the experiment that the proposed hue-invariant FID
is more consistent with human preferences than the original FID and could serve as an appropriate evaluation metric for image colorization task.

Our contributions can be summarized as follows:
\begin{itemize}
    \item To the best of our knowledge, we are the first to comprehensively review and benchmark language-based image colorization literature on automatic and language-based image colorization tasks. We present insightful analysis on problem formulations, model architectures, and training schemes, and discuss future directions.
    
    \item We explore the generation power of diffusion models and conduct a deeper discussion on different condition insertion methods with comparative experiments. We reveal that different insertion methods provide different feature granularity and intensity, leading to different limitations respectively. We believe such comprehensive and in-depth analysis can provide more valuable insights to the community.
    \item Motivated by the limitations revealed by our survey, we fully explore the generative potential of diffusion models and propose a simple yet effective colorfulness controllable colorization method \textbf{Color-Turbo}, which can produce more stable colorized results and accelerate the inference by 14 times. 
    
    \item We propose a new metric hue-invariant FID, which is more consistent with human preferences. Based on the proposed evaluation protocol, we benchmark representative language-based methods on three evaluation datasets and analyze the impacts of different training schemes and model architecture systematically, shedding light on future research directions.
    
\end{itemize}

The remaining sections are organized as follows: We first present a comprehensive review of automatic and language-based image colorization methods 
in Sec.~\ref{sec:review_uncond} and Sec.~\ref{sec:review_cond}, respectively. Subsequently, based on our analysis, we propose Color-Turbo to achieve  colorfulness controllable and efficient language-based colorization in Sec.~\ref{sec:color-turbo}. Then, we comprehensively benchmark previous methods, and demonstrate the superiority of our proposed method in Sec.~\ref{sec:benchmark}. Finally, we conclude and summarize with possible future directions in Sec.~\ref{sec:conclusion}.

\vspace{-3mm}
\section{Review on Automatic Image Colorization}
\begin{figure*}
    \centering
    \includegraphics[width=\linewidth]{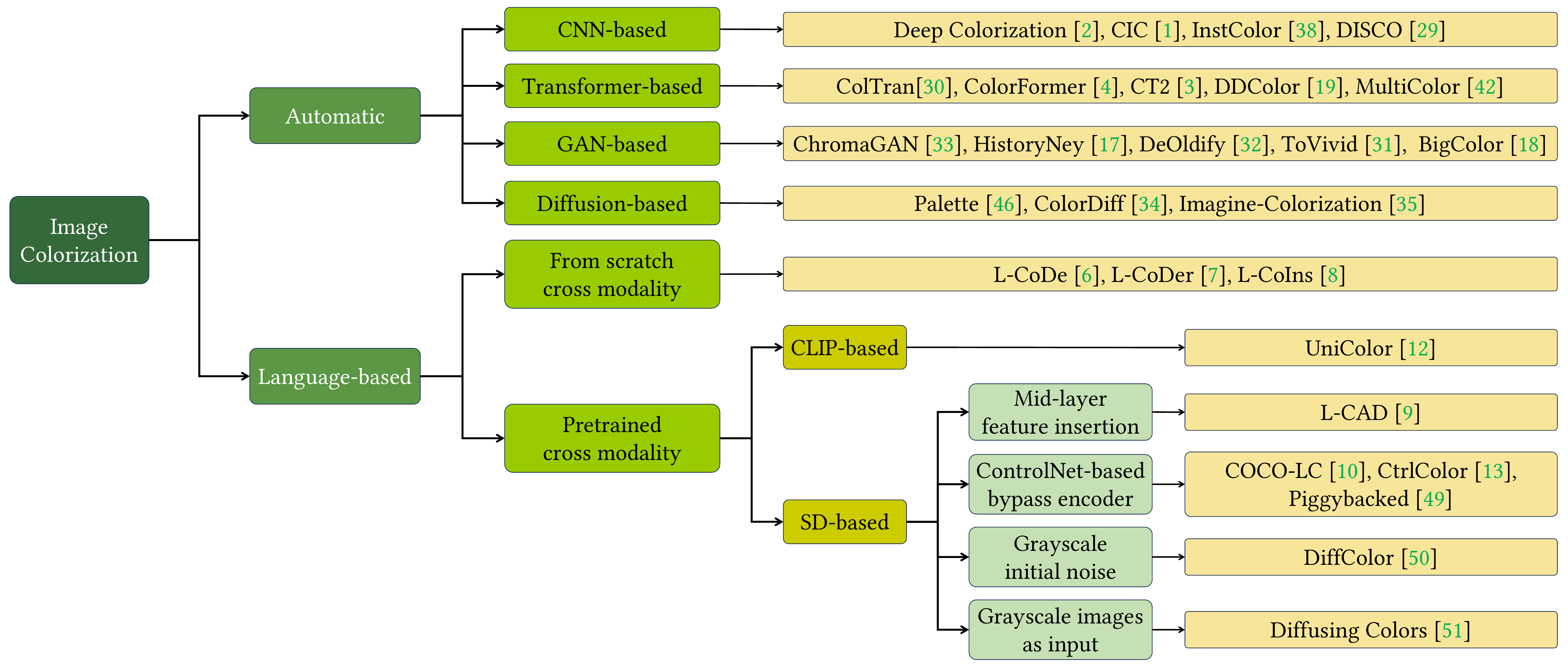}
    \caption{\ys{Taxonomy} of automatic and language-based methods.}
    \vspace{-4mm}
    \label{survey}
\end{figure*}
\label{sec:review_uncond}

In this section, we will discuss automatic image colorization methods.
In this domain, researchers have made a lot of efforts on exploring different model architectures such as CNNs and Transformer to fit natural image distribution accurately~\cite{cic, DISCO, HistoryNet, coltran, ct2, colorformer, ddcolor}. Meanwhile, the task can be solved as a conditional generation problem with grayscale images as conditions~\cite{tovivid, DeOldify, ChromaGAN, bigcolor, colordiff, imagination}, by exploiting the powerful generative priors of GAN~\cite{gan} and Diffusion models~\cite{ddpm, sde}. Therefore, based on the base models, automatic image colorization methods can be roughly devided as CNN-based, Transformer-based, GAN-based and Diffusion-based methods.

\vspace{-3mm}
\subsection{CNN-based Methods}
\label{cic}
With the development of deep learning, the most straightforward way of image colorization is to fit a Convolutional Neural Network~(CNN) to fit the color image distribution.
Cheng \etal~\cite{deep_color} proposed the first deep neural network for image colorization, but limited by over-smoothing regression loss and insufficient datasets.
To mitigate the issues of under saturated and plain colors, CIC~\cite{cic} formulates colorization as a classification problem in quantized CIELAB color space with cross-entropy loss.
InstColor~\cite{instColor} introduces bounding boxes and colorizes each crop-out instance respectively with a fusion module to merge each instances smoothly. However, the incorrect external prior extraction may instead aggravate color overflows.
DISCO~\cite{DISCO} designs a coarse-to-fine colorization framework, which colorizes anchor points firstly, then predicts the remaining pixel colors by referring the sampled anchor colors. 
With such disentangled framework, the model can learn global affinity between the anchors and other primitives, reducing color ambiguity. 

\vspace{-3mm}
\subsection{Transformer-based Methods}
Transformer~\cite{transformer} and  ViT~\cite{vit} has developed rapidly with its superb ability on long-range modeling.
ColTran~\cite{coltran} proposes the first transformer-based colorization model incorporates a multi-stage colorization strategy.
Nonetheless, ColTran hampers its ability with limited inductive bias as CNN.
ColorFormer~\cite{colorformer} proposes a Global-Local hybrid self-attention to balance and a color memory which stored multiple groups of semantic-color mapping for efficient color acquisition.    
CT2~\cite{ct2} computes statistics of color space and conduct 313 meaningful color tokens. On the basis of this, CT2 leverages an adaptive attention mechanism to bridge correspondence between such prior color tokens and luminance tokens.
Inspired by query-based ViT~\cite{detr}, DDColor~\cite{ddcolor} presents a set of learnable color tokens and a cross attention-based fusion module to merge the grayscale and color features.
MultiColor~\cite{MultiColor} builds upon DDColor~\cite{ddcolor} and proposes a multi-branch model, fully leveraging complementarity and more nuanced color information based on the inconsistent color space.

\vspace{-2mm}
\subsection{GAN-based Methods}
GANs have achieved remarkable success in synthesizing high-fidelity natural images.
ChromaGAN~\cite{ChromaGAN} simply leverage a typical PatchGAN~\cite{patchgan} discriminator with WGAN~\cite{wgan} loss to maintain realism.
HistoryNet~\cite{HistoryNet} incorporates classification and segmentation information with according modules and losses. HistoryNet also collects a large-scale dataset from old movies.
Compared with traditional GAN training, Deoldify~\cite{DeOldify} proposes an efficient GAN training strategy that adopts an asynchronous training mode on generator and discriminator. 
ToVivid~\cite{tovivid} uses the generative prior of pre-trained BigGAN~\cite{biggan} in an inversion manner.
However, GAN inversion is time-consuming and inaccurate especially on grayscale images. The mismatches between GAN-inverted images and input images are inevitable, leading to color distortions.
In view of above drawbacks, BigColor~\cite{bigcolor} builds an encoder-generator model, which equipped with pre-trained BigGAN's generator and discriminator. 
Unfortunately, it requires a high training cost, and may produce color artifacts as shown in Fig.~\ref{discuss_uncond}(c).

\vspace{-3mm}
\subsection{Diffusion-based Methods}
\label{sec:uncond_diff}
Diffusion models~\cite{ddpm, sde, sd, sd-turbo} can produce realistic images through an iterative denoising process.
However, as diffusion sampling process is stochastic, it is hard to maintain both structural consistency and generative ability.
Palette~\cite{saharia2022palette} seizes the powerful fitting ability of diffusion paradigm and trains a diffusion model from scratch.
ColorDiff~\cite{colordiff} uses segmentation maps and designs a high-level semantic insertion module based on~\cite{sd}. ColorDiff also builds skip connections between VAE encoder and decoder to maintain structure consistency.
Cong \etal~\cite{imagination} proposed a ControlNet-based colorization framework with coarse-to-fine design, where coarse colorized images are produced by pre-trained ControlNet~\cite{controlnet} with edge maps and finer results are generated by Semantic-SAM~\cite{semantic_sam} and a multi-level hint points propagation strategy.
However, multiple coarse images generation is time-consuming. In addition, the incorrect segmentation results, especially inferred from grayscale images, lead to potential color overflows.

\vspace{-3mm}
\subsection{Discussion}
Although automatic colorization methods have made a lot of progress, the above four categories of methods still have their own problems, such as under saturation, color overflow, color artifacts, color incompleteness and color confusion, as shown in Fig.~\ref{discuss_uncond}.

Among CNN-based methods, some better training strategies are designed, including better loss function~\cite{cic}, decoupled two-stage~\cite{DISCO} and external information involved~\cite{instColor}. However, due to the color ambiguity, these methods based on regular regression or classification losses suffer from grayish results and color overflows.

\ys{Compared to CNN-based methods, Transformer-based methods are capable of building long-range visual dependency with flexible attention mechanisms, which is however computational inefficient.
To mitigate such limitation, GL-MSA~\cite{colorformer} and query-based transformer~\cite{ddcolor, MultiColor} are proposed.
However, these methods are still limited by the common color confusion problem due to color ambiguity, such as the red and blue sign in Fig.~\ref{discuss_uncond}(e).}

\ys{GAN-based methods either train from scratch~\cite{ChromaGAN, DeOldify}, which still produces under saturated results and color overflows, or utilize pre-trained GAN priors~\cite{tovivid, bigcolor}, which suffers from inaccurate GAN inversion or unstable training. Although the inference of GAN is efficient compared to Transformer, the training of GAN is well-known to be unstable, leading to color artifacts as shown in Fig.~\ref{discuss_uncond}(c).}

Diffusion-based methods suffer the inherent long inference time of diffusion process. Moreover, training a diffusion model from scratch~\cite{saharia2022palette} generally has poor generalization due to the limited dataset, while utilizing pre-trained diffusion models with external priors~\cite{colordiff, imagination} may affected by inaccurate priors extracted from grayscale images.

Overall, the shared key challenge of automatic colorization is the inevitable color ambiguity. By comparison, language-based colorization provide sufficient user controllability, as shown in Fig.~\ref{compare_uncond_cond}. Therefore, this paper focuses on language-based methods and expand a comprehensive review in Sec.~\ref{sec:review_cond}.

\vspace{-4mm}
\section{Review on Language-based Image Colorization}
\label{sec:review_cond}

Image colorization is inherently an ill-posed problem with multiple reasonable choices. Language-based image colorization methods reduce such uncertainty through user-provided instructions in a natural and flexible way, which enhances controllability. 
Apart from the basic issues mentioned above, 
the key point of language-based methods lies on how to establish an accurate cross-modal correspondence between color words in text prompts and objects in grayscale images. According to the way of building cross-modal correspondence, language-based image colorization methods can be divided into two categories according to whether they use pre-trained cross-modal models. We further summarize their main features in Table~\ref{main_features_sum} briefly.
\begin{table*}[h]
\label{main_features_sum}
\caption{Summary of key features of language-based image colorization methods.}
\resizebox{\linewidth}{!}{
    \begin{tabular}{c|c|c}
    \toprule
    \multicolumn{1}{c|}{Category} & \multicolumn{1}{c|}{Method} & \multicolumn{1}{c}{Main features} \\
    \midrule
    \multirow{3}{*}{\thead{From scratch \\cross-modality}} & L-CoDe~\cite{lcode} & Hand-annotated color-object decouping with BiLSTM for texts and CNN for images \\
     & L-CoDer~\cite{lcoder}  & Hand-annotated color-object decouping with Transformer and mixed color-object tokens \\
     & L-CoIns~\cite{lcoins} &  Adatpive cross-modal ability with Group Transformer, use color augmentation to cut statistic bridge between luminance and color\\ 
     \midrule
    \multirow{7}{*}{\thead{Pretrained \\cross-modality}} & UniColor~\cite{unicolor} & Use CLIP to bridge language and grayscale images, use autogressive Transformer to generate images with adversarial training \\
     & L-CAD~\cite{lcad} &  Finetune Stable Diffusion to align cross modality and insert luminance conditions into VAE decoder\\
     & piggybacked-color~\cite{piggybacked} &  Similar to ControlNet, but finetunes the whole U-Net instead   of only encoder \\
     & DiffColor~\cite{diffcolor} &  Grayscale noisy latents as input through DDPM forward, color contrastive loss, multi-stage colorization\\
     & CtrlColor~\cite{CtrlColor} & Use ControlNet to finetune Stable Diffusion \\
     & Diffusing Colors~\cite{Diffusing-Colors} & Reformulate denoising loop to recoloring process with increasing saturation and adopt DDIM as sampling strategy\\
     & COCO-LC~\cite{coco-lc} & Coarse-to-fine framework based on ControlNet and Stable Diffusion with multi-level cues, colorfulness controllable decoder \\
     \bottomrule
    \end{tabular}
}
\vspace{-3mm}
\end{table*}

\vspace{-3mm}
\subsection{Train a Cross-modality Model from Scratch}
\begin{figure}
    \centering
    \includegraphics[width=1\linewidth]{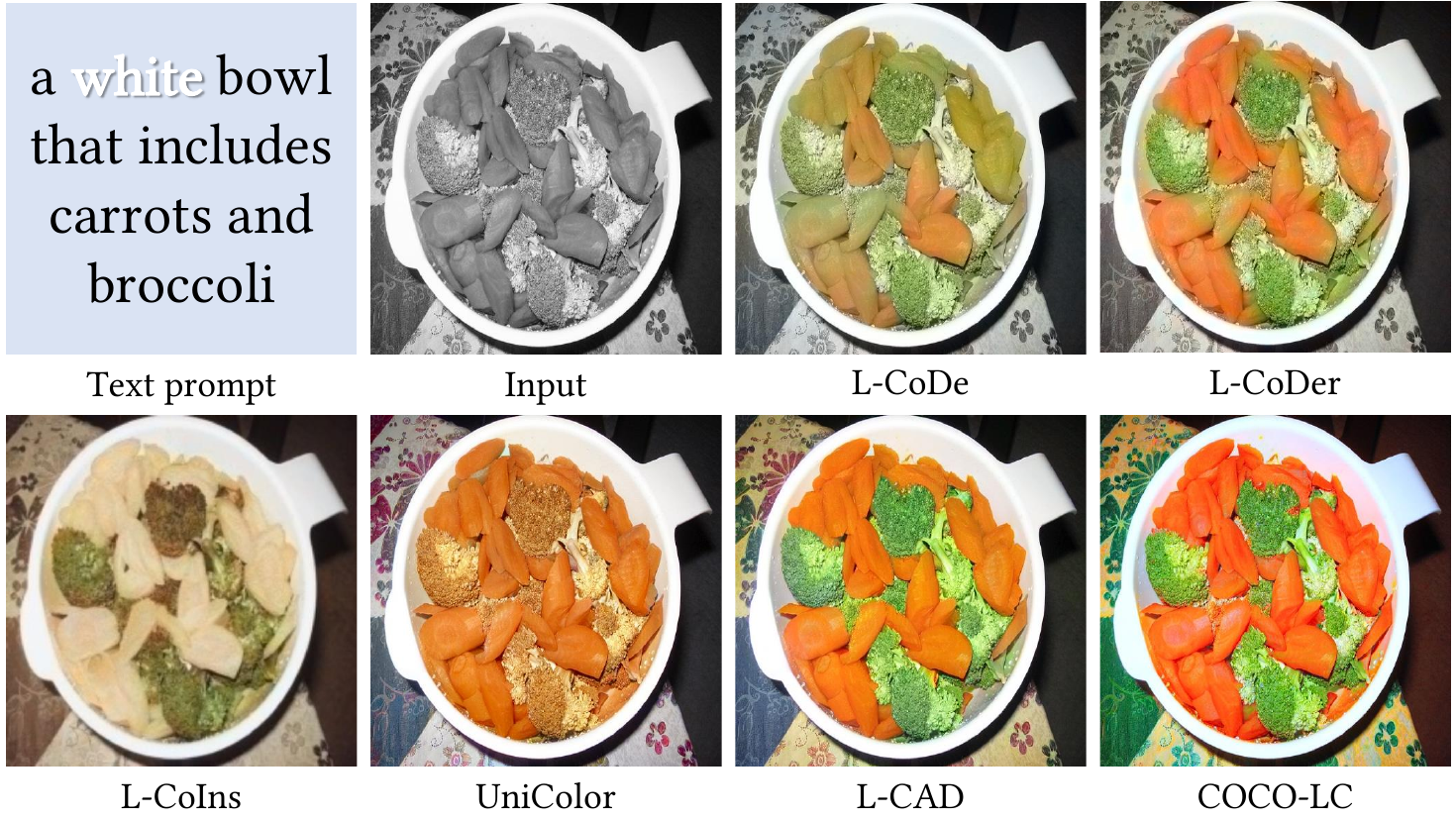}
    \caption{\lyf{Illustration of main features of existing language-based methods. Since L-CoDe~\cite{lcode} and L-CoDer~\cite{lcoder} decouple color-object words, color overflows will occur on some objects which have no color words accordingly. L-CoIns~\cite{lcoins} weakens the correlation between brightness and color with data augmentation, damaging the original semantics, as carrots are turned into yams. UniColor~\cite{unicolor} cannot achieve finer colorization, as broccoli turns to red. Diffusion-based L-CAD and COCO-LC produce more plausible results. COCO-LC generate more colorful image, but with some color overflows on the bowls compare with L-CAD.}}
    \vspace{-6mm}
    \label{lcoins_fails}
\end{figure}

Before the emergency of large-scale cross-modal models~\cite{clip, sd}, several attempts have been made to construct a cross-modality model for colorization from scratch.
\begin{itemize}
    \item \textit{L-CoDe:} To address issues of inaccurate alignment between color words and gray images, L-CoDe~\cite{lcode} locates the problem to color-object coupling and color-object mismatch, aiming to find exactly the corresponding area in the images for each word in the text prompts. To provide additional supervised signals, L-CoDe collects a dataset with annotated object nouns and color adjectives pairs in each text prompt. With such relationship in text prompts, colorization models can transfer the correspondence between visual regions and nouns to the correspondence between regions and adjectives, leading to text-aligned colorized results. However, L-CoDe produces unrealistic results due to simple feature extractor and fusion module.
    \item \textit{L-CoDer:} On the basis of L-CoDe, L-CoDer~\cite{lcoder} insists the idea of using additional supervision to find more precise correspondence between colors words and objects nouns, and makes progress on two aspects. \lyf{Considering the poor model architecture of L-CoDe}, L-CoDer adopts an advanced Transformer-based model, treating image features and text features both as tokens in a unified way, which narrows the domain gaps between visual and language modalities. As for the poor ability of text feature extractor, L-CoDer utilizes a pre-trained language model, BERT~\cite{bert}, as its language encoder. Although L-CoDer produces more plausible results compared to L-CoDe, \lyf{it still struggles to bridge visual-language domain gap well with limited dataset scale.}
    
    \item \textit{L-CoIns:}  
    Different from explicitly learning a manually annotated correspondence between color words and objects nouns, L-CoIns~\cite{lcoins} considers an image as a composition of a number of groups with similar colors, hence adopting a grouping mechanism to aggregate similar image patches for identifying corresponding regions to be colorized, which flexibly extracts and fuses image and text features. 
    Besides, L-CoIns presents a novel data augmentation strategy to cut down the statistical correlation between luminance and color, so that L-CoIns can produce colorization results that are more consistent with the given language description. However, the lower correlation between brightness and color may damages visual appearance under limited language understanding ability.  
    For example, in Fig~\ref{lcoins_fails}, it mistakenly colorizes the dark carrots to light yams.
\end{itemize}

Although L-CoDe, L-CoDer and L-CoIns all make efforts to find an accurate correspondence between color words and gray images, none of them exploit the powerful cross-modality models pre-trained on large-scale dataset, leading to color distortions and poor generalization.

\vspace{-3mm}
\subsection{Utilize a Pretrained Cross-modality Model}
\subsubsection{Use CLIP to align text and image}
CLIP~\cite{clip} has shown its superior ability to align language domain and visual domain. UniColor~\cite{unicolor} is one of the first methods to introduce this powerful cross-modality model into language-based image colorization task. UniColor converts text descriptions to a set of hint points which indicate specific colors of grid regions with the help of CLIP. It then uses these hint points as constraints to make the colorized images consistent with text prompts. Specifically, UniColor first extracts color words and corresponding object words in each text prompt, and divides the whole gray image into grids with a fixed size. Then UniColor uses a sliding window browsing all grids to compute correspondence of each grid and object words, measured by the cosine similarity between CLIP embeddings. After that, UniColor broadcasts such relationship between each grid and object words to the correspondence between each grid and color words under the instruction of (color word, object word) pairs, leading to hint points with required colors.
Based on the hint points, UniColor constructs a hybrid-Transformer and a Chroma-VQGAN to generate final colorized results. The hybrid-Transformer predicts color features in an autoregressive manner with the compensation of hint points. Then the color features are fed into the decoder of Chroma-VQGAN, which is trained by an adversarial loss to achieve realistic results.

However, this method is limited by the grid size and stride of sliding windows, hard to achieve finer colorization spatially. As shown in Fig.~\ref{lcoins_fails}, UniColor produces more color overflows, leading to irregular ``red broccoli".

\subsubsection{Exploit Stable Diffusion for colorization}
\label{sec:cond_sd}
As the great physicist Feynman said, ``What I cannot create, I do not understand''. Thanks to the superior performance of Stable Diffusion on text-to-image generation, there is no doubt on its capability to align language domain and visual domain, making it a more effective paradigm of language-based image colorization than previous works. 
\ys{Since Stable Diffusion is originally designed for text-to-image generation, when applying it to colorization, it is an essential issue on how to insert meaningful semantic and structure information of grayscale images into the generation process.}

\begin{figure*}[h]
    \centering
    \includegraphics[width=0.9\linewidth]{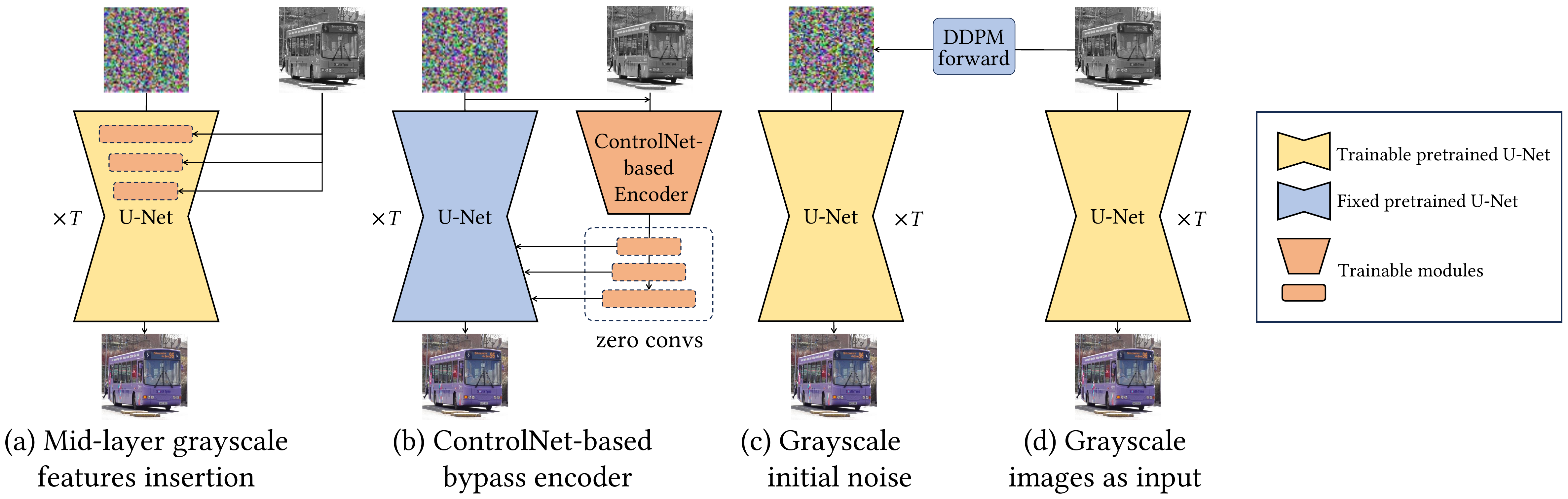}\vspace{-3mm}
    \caption{We summarize four representative condition insertion paradigms, including (1) mid-layer grayscale features insertion, (2) ControlNet-based bypass encoder, (3) grayscale initial noise, (4) grayscale images as input. These four methods can constraint the generation process by grayscale images with different injection intensities and feature granularity.}
    \vspace{-6mm}
    \label{conditions}
\end{figure*}

We summarize four typical condition insertion paradigms: (1) mid-layer grayscale features insertion, (2) ControlNet-based bypass encoder, (3) grayscale initial noise, (4) grayscale images as input, as illustrated in Fig.~\ref{conditions}.

\textbf{(1) Mid-layer grayscale features insertion.} L-CAD~\cite{lcad} proposes channel extended convolutional blocks (CEC blocks) to insert luminance features into middle layers of U-Net, providing essential structure and content information. Those CEC blocks are initialized to zero to maintain training stability. Besides, to expedite training convergence, CEC blocks are only added to the encoder part of U-Net.

Furthermore, to maintain structure consistency between colorized results and grayscale inputs, L-CAD presents a luminance-aware decoder, which adds skip connections between the VAE encoder and decoder, similar as ColorDiff~\cite{colordiff} mentioned at Sec.~\ref{sec:review_uncond}. L-CAD also proposes a training-free instance-aware strategy to achieve precise colorization with users' annotations of particular object masks and their corresponding color word requirements. Specifically, L-CAD assigns user-specified color words to certain masks by modulating the cross attention maps in the U-Net of Stable Diffusion through iterative optimization.

\textbf{(2) ControlNet-based bypass encoder.} 
ControlNet~\cite{controlnet} utilizes a trainable copy of U-Net encoder as the \textit{condition branch} to extract semantic features of condition images and insert condition features into U-Net decoder's layers through zero-initialized convolution blocks, which maximally preserve the generative potential of \textit{diffusion branch}.
Directly utilizing a ControlNet on image colorization task with grayscale images as conditions is a straightforward idea.
COCO-LC~\cite{coco-lc} and CtrlColor~\cite{CtrlColor} both adopt ControlNet architecture, while COCO-LC leverages more meaningful high-level and low-level cues (semantic segmentation maps or edge maps extracted by SAM~\cite{sam}) to further reduce color overflows and artifacts.
Notably, COCO-LC presents a novel coarse-to-fine colorization framework to alleviate inaccurate correspondence between the color words and the spatial regions in grayscale images. Specifically, the cross-modality alignment ability of cross attention mechanism of Stable Diffusion is weakened on gray images. COCO-LC observes such obstacle and designs a Color Information Adapter (CIA) after the VAE Encoder, conducting a coarse colorization with the help of color features extracted from color images or color text prompts by CLIP, which are effective to build a precise mapping between color words and gray images with less color overflow.

Likewise, Liu \etal~\cite{piggybacked} proposed a piggybacked diffusion models. This method builds a Diffusion Guider with the same architecture of U-Net in Stable Diffusion, to extract and inject feature maps into the U-Net in Stable Diffusion.
However, this method utilizes both the U-Net encoder and decoder as condition branch, which doubles the trainable parameters with much more training resources.

\begin{figure}
    \centering
    \includegraphics[width=1\linewidth]{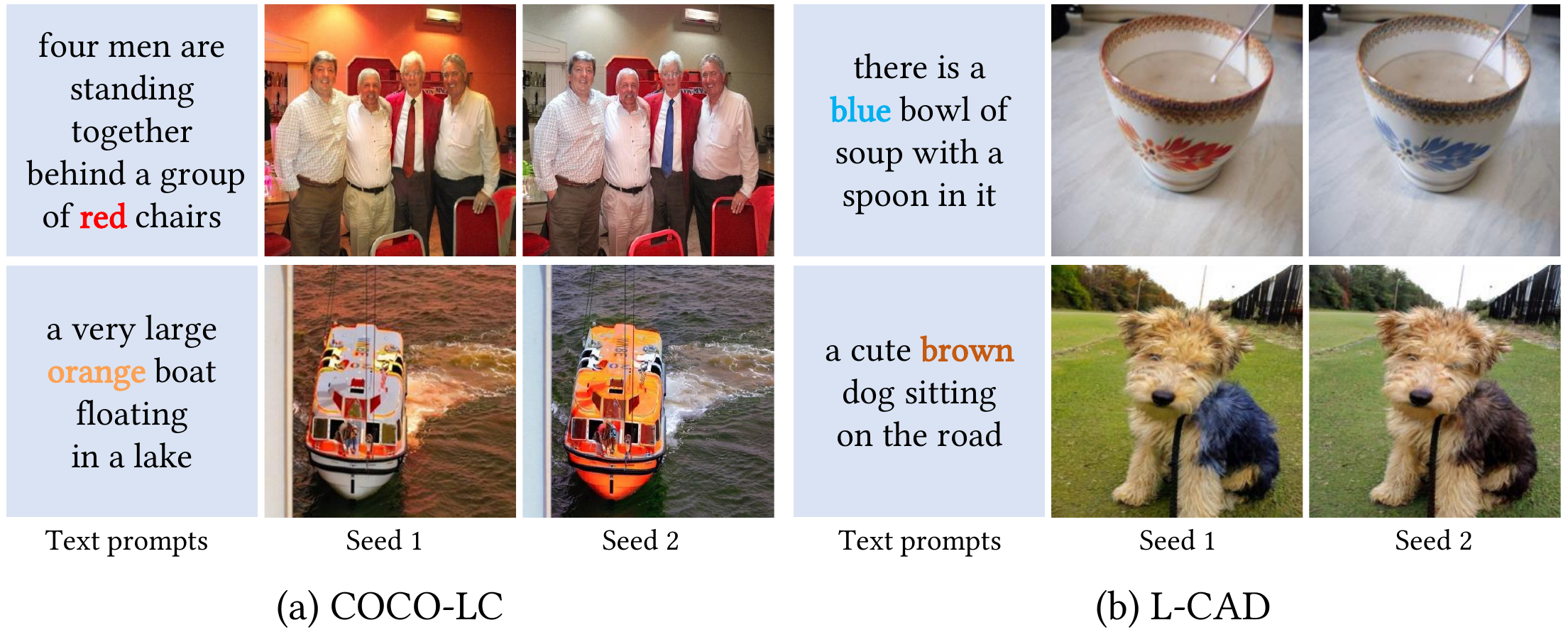}\vspace{-3mm}
    \caption{Qualitative results of COCO-LC~\cite{coco-lc} and L-CAD~\cite{lcad} with different initial noises. We use different random seeds to get different Gaussian noises. The initial noises yield huge difference in the output.}
    \vspace{-6mm}
    \label{diff_seed}
\end{figure}

Although ControlNet provides an effective paradigm to insert new conditions, the noisy features in diffusion branch and the content features in condition branch may cause feature conflicts~\cite{img2img-turbo}, leading to unstable colorization under different initial noise, as shown in Fig.~\ref{diff_seed} (a).

\textbf{(3) Grayscale initial noise.} DiffColor~\cite{diffcolor} presents another way of inserting grayscale images as conditions during generation process. After obtaining the latent code $z_0$ of grayscale images by VAE Encoder, DiffColor adopts DDPM forward process to add noises on $z_0$, resulting in $z_T$, as the initial input of the denoising network. On the basis of this, DiffColor finetunes the denoising network with a contrastive loss based on CLIP text encoder to constrain colorization with text prompts more accurately and proposes a multi-stage colorization framework to further achieve accurate colorization under text prompts.

\textbf{(4) Grayscale images as input.} The aforementioned methods take Gaussian noise or noisy latent codes as input, which leads to inherent stochastic and unstable generation. Diffusing Colors~\cite{Diffusing-Colors} provides a novel perspective to utilize the strong generative priors and cross-modality ability of Stable Diffusion. Inspired by Cold Diffusion~\cite{ColdDiffusion}, it finetunes the denoising Stable Diffusion to a network that progressively enhances colorfulness. We have explained more details of it in the Supplementary Material. 

\textbf{Discussion.} The above four paradigms steer the generation process of colorized images towards the structure of the grayscale images and the provided text descriptions. However, their respective injection intensities and feature granularity lead to different trade-off between color generation and structural consistency.

\begin{itemize}
    \item \textbf{Mid-layer grayscale features insertion.} 
    Similar to ControlNet-based methods, mid-layer grayscale features insertion guides the colorization process with the features of grayscale images. The different is that methods like L-CAD~\cite{lcad} utilize VAE encoder as feature extractor, which pays more attention on local details information in a finer granularity compared with U-Net encoder. Moreover, L-CAD proposes to insert grayscale features into layers of U-Net encoder with CEC blocks, leading to more stable colorization. 
    However, such insertion degrades the generative ability of original Stable Diffusion, making the method hard to colorize images beyond natural images, such as artistic images. 
    \item \textbf{ControlNet-based bypass encoder.} ControlNet-based methods~\cite{coco-lc, CtrlColor} decouple the colorization model into generation process and condition insertion.
    However, the low-resolution feature space of the encoder of U-Net is limited to provide sufficient local structure and details information of grayscale images, leading to structural distortion and uncontrollable artifacts during the colorization process.
    
    \item \textbf{Grayscale initial noise.} In addition to using the intermediate features as conditions, some methods~\cite{diffcolor, colordiff} propose to use the initial noisy latent inferred from the grayscale images to provide conditions. Nonetheless, it is hard for the low-resolution noisy latents to provide sufficient control, \lyf{hampering their ability to steer the diffusion generation process towards proper direction}. 
    
    \item \textbf{Grayscale images as input.} Using the high-resolution grayscale images as the inputs, Diffusing Colors~\cite{Diffusing-Colors} combines the generation process and condition injection effectively with a reformed colorfulness enhancement progress. 
    However, since this type of methods abolish initial noises and denoising process of origin diffusion models, they consume more training resources and require a larger amount of data to ensure that the model can converge compared to other diffusion-based fine-tuning methods. Moreover, this type of methods are hard to preserve structures well because they have to train models under multiple grayscale levels, leading to accumulative errors.
\end{itemize}

Considering the above attempts previous methods have made, we argue that it is still hard to achieve a good balance between color generation and structural consistency. The first two paradigms utilize mid-layers features as conditions to guide color generation, which are lack of detailed local information, and may lead to color overflows and artifacts. The last two paradigms insert conditions in a more direct way through changing input, but may be limited by insufficient control or more training cost.
Moreover, the above diffusion-based methods requires an iterative and time-consuming denoising process, and some methods are fragile to random initial noises as shown in Fig.~\ref{diff_seed}.
\ys{Therefore, a more efficient and effective method is desired. 
In view of the above, we are motivated to investigate a new language-based image colorization framework that achieves a better trade-off between generative ability and condition consistency, which is detailed in Sec.~\ref{sec:color-turbo}.}

\vspace{-4mm}
\section{The proposed method: Color-Turbo}
\label{sec:color-turbo}
\vspace{-1mm}
\subsection{Motivation}
Automatic image colorization methods do not provide enough control for users and therefore cannot meet the needs of practical applications. By comparison, text description is the most natural and easy-to-use way compared with other external instructions like hint points, color strikes or color palettes, \ys{rising the research on language-based image colorization.}
Although recent language-based methods leveraging the powerful diffusion models demonstrate satisfying performance, they are highly sensitive to the initial noise, leading to unstable performance, as shown in Fig.~\ref{diff_seed}, a phenomenon also demonstrated by~\cite{optimalBC, init1, init2}. 

%
Apart from the above over-sensitiveness, diffusion-based methods are also time-consuming, which greatly impedes their application on mobile devices and potential extension on video colorization.  
Furthermore, different methods insert grayscale conditions with different feature granularity and intensity, struggling to achieve an optimal balance between color generation and structure consistency.

Based on the above analysis, we would like to design a new baseline.
First, we expect the new baseline to produce plausible colorized results with the guidance of flexible text prompts. Thus, we adopt Stable Diffusion~\cite{sd} as the backbone with its powerful cross-modal ability and generation performance, formulating colorization as a conditional generation problem.
Second, we would like to improve efficiency.
Inspired by img2img-turbo~\cite{img2img-turbo}, we implement a distilled diffusion model to predict the denoised results in a single step, which significantly accelerates the inference.
\ys{Third, to avoid the aforementioned over-sensitiveness to stochastic initial noises, we take grayscale images as input instead of Gaussian noises, which essentially solve the problem.} Additionally, based on the analysis in Sec.~\ref{sec:cond_sd}, we would like to achieve a better balance between color generation and structural consistency. 
We construct an end-to-end \ys{single-step} model and jointly train VAE and U-Net, which prevents error accumulation caused by iterative inference as in~\cite{Diffusing-Colors} and takes both structural and semantic features into consideration.
Furthermore, to maintain structure information precisely and control output colorfulness, we construct skip connections between VAE encoder and decoder with a flexible scale factor.
Our method directly utilizes grayscale images as input to a single-step end-to-end model, which outperforms other skip connection-based methods~\cite{piggybacked, lcad, colordiff} in structure preservation.



In view of the above, we propose our method, \textbf{Color-Turbo}, a distilled one-step diffusion model for efficient colorfulness controllable language-based image colorization. 
We shall introduce more details as follows.

\vspace{-4mm}
\subsection{Color-Turbo for colorfulness controllable and efficient language-based image colorization}
\begin{figure}[h]
    \centering
    \includegraphics[width=\linewidth]{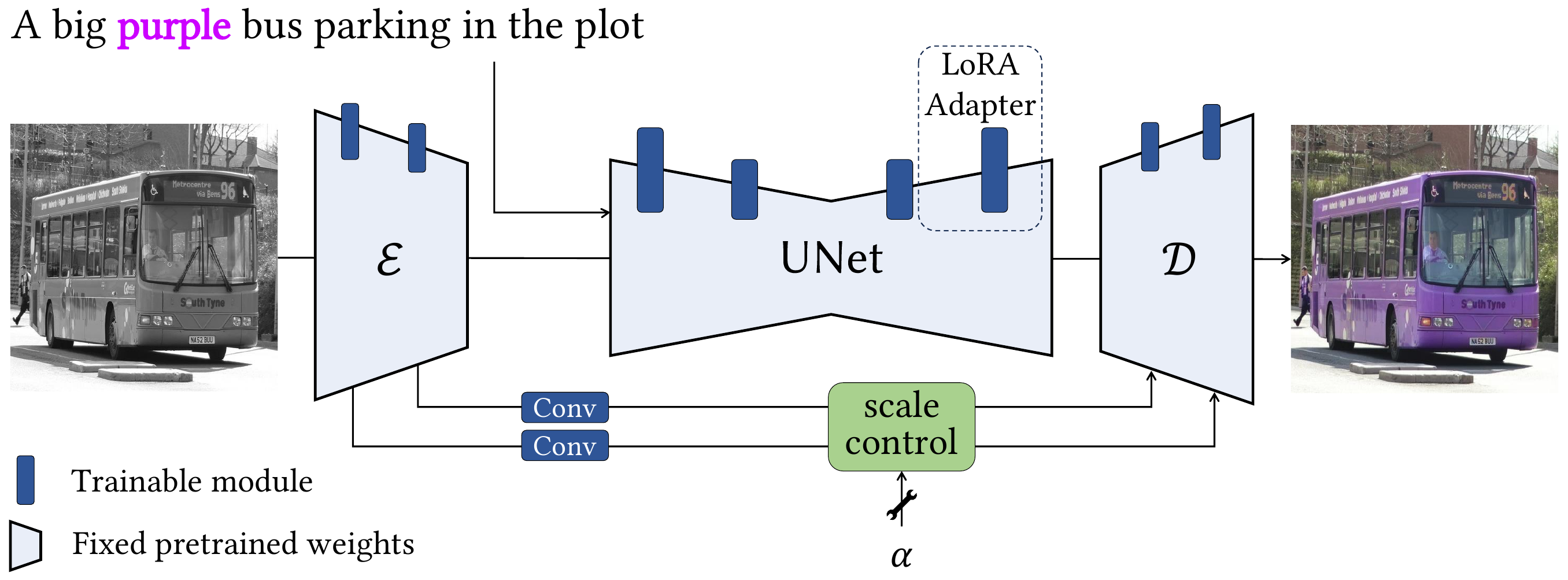}
    \caption{Overview of our proposed Color-Turbo framework. Our method achieves efficient and colorfulness controllable language-based image colorization. We initialize the U-Net and VAE with the pre-trained Stable Diffusion model, and only train LoRA adapters to finetune the model on colorization task. To maintain the luminance consistency, we add skip connections between VAE encoder and decoder with zero-init convolution blocks and a flexible scale factor $\alpha$, further achieving diverse colorized results with different colorfulness.}
    \vspace{-3mm}
    \label{Color-Turbo-pipe}
\end{figure}

The proposed Color-Turbo is a single step image-to-image translation framework based on the pre-trained Stable Diffusion model and LoRA adapters~\cite{lora}, as shown in Fig.~\ref{Color-Turbo-pipe}.
\vspace{-4mm}

Specifically, we adopt the model architecture of img2img-turbo~\cite{img2img-turbo}. 
We utilize LoRA adapters to finetune pretrained VAE and U-Net of Stable Diffusion model.
To maintain luminance structure consistency, skip connections between VAE Encoder and Decoder is added, with a 1$\times$1 convolutional layer in the middle. 
To achieve diverse results with different colorfulness smoothly, we add a control factor $\alpha$ on the skip connections, to scale the grayscale features' intensity flexibly. 
Lower $\alpha$ brings more colorful results. More visualized results are shown in Fig.~\ref{fig:ablation_alpha}.

As for the training objectives, we utilize L2 distance in Euclid space and VGG~\cite{vgg} feature space between ground truth color images and colorized images as basic regression loss, denoted by $\mathcal{L}_{pixel}$ and $\mathcal{L}_{lpips}$ respectively. 
We further use CLIP loss 
\begin{equation}
    \label{clip_loss}
    \mathcal{L}_{clip} = 1 - \mathcal{M}(\mathcal{F}(I_{pred}), \mathcal{F}(\mathcal{P})),
\end{equation}
to ensure color accuracy constrained by text prompts, where $I_{pred}$ is the colorized image, $\mathcal{P}$ is the text description, $\mathcal{F}$ is the CLIP model, and $\mathcal{M}$ denotes the cosine similarity measurement.
Moreover, inspired by~\cite{aided-gan}, we utilize an adversarial loss 
$\mathcal{L}_{adv}$ on the basis of CLIP image encoder, achieving more realistic visual performance and maintain the training stability at the same time. The overall loss function can be formulated as
\begin{equation}
    \begin{aligned}
        \label{Color-Turbo-loss}
        \mathcal{L} & = \lambda_{pixel} \cdot \mathcal{L}_{pixel} + \lambda_{lpips} \cdot \mathcal{L}_{lpips} \\ &
        + \lambda_{clip} \cdot \mathcal{L}_{clip} + \lambda_{adv} \cdot \mathcal{L}_{adv}.        
    \end{aligned}
\end{equation}

\vspace{-4mm}
\subsection{Strengths}
Color-Turbo is a simple yet effective baseline for \lyf{colorfulness controllable and efficient} language-based image colorization.
\lyf{Firstly,} to satisfy diverse requirments of different users, Color-Turbo controls colorfulness of results by a simple scale factor $\alpha$ flexibly, broaden its practical applications. Besides, Color-Turbo is a kind of distilled diffusion model, which directly convert a grayscale image input to a colorful images in a single denoising step, 14 times faster than existing diffusion-based colorization methods~\cite{coco-lc, lcad}, significantly reducing the inference cost. As a result, Color-Turbo can be applied to video colorization directly.

Moreover, Color-Turbo does not take random noises as input, maintaining inherent stability compared with previous diffusion-based colorization methods. As some works (\cite{optimalBC, init1, init2}) have indicated that different initial noise may lead to dramatically different results, previous diffusion-based methods will limited to such over-sensitive property, our Color-Turbo can avoid this problem in the essence. 
Furthermore, since Color-Turbo does not depend on any initial random noises that may amplify subtle variations across consecutive frames and hinder consistent video application, it can generate more coherent colorized video, sheding more light on the colorization of monochrome films.

\vspace{-3mm}
\section{Experimental Results}
\label{sec:benchmark}
\subsection{Evaluation Dataset}
\begin{itemize}
    \item \textbf{Extended COCO-Stuff Dataset} is built upon COCO-Stuff dataset~\cite{coco-stuff} by discarding unqualified samples for the colorization task and is first collected by L-CoDe~\cite{lcode}. We further filter out some black and white photos, remains 1,946 images with 3,520 text prompts.
    \item \textbf{Multi-instances Dataset} includes multiple instances with different visual characteristics within a single image, including 7,213 images with 12,709 text prompts. The text description of this dataset contains more detailed descriptions of objects and corresponding colors than Extended COCO-Stuff Dataset. This dataset is first collected by L-CoIns~\cite{lcoins}.
    \item \textbf{ImageNet-val5k}. It is a common practice in colorization field to use the first 5,000 images in ImageNet~\cite{imagenet} as evaluation dataset. We use BLIP~\cite{blip} to generate a default text description of the color image to evaluate language-based methods. 
\end{itemize} 

\vspace{-5mm}
\subsection{Evaluation Metrics}
\begin{itemize}
    \item \textbf{FID}~\cite{fid} measures the overall quality of colorized images without paired data, which is suitable for ill-posed colorization problem.
    \item \textbf{Hue-invariant FID \lyf{(HI-FID)}.} We propose this metric based on FID, which ignores the difference of color hues and concentrates more on color overflow or color artifacts. We will introduce its details at Sec.~\ref{fid_correct}.
    \item \textbf{Colorfulness \lyf{(CF)}} is proposed by~\cite{hasler2003measuring} to evaluate the vividness of images, and is close to human perception.
    \item \textbf{$\Delta$Colorfulness \lyf{($\Delta$CF)}} computes the discrepancy of colorfulness between ground truth and colorized images. 
    \item \textbf{CLIP Score} evaluates \ys{how well the results match the text prompts for} language-based colorization methods, which computes cosine similarity of image embedding and text embedding extracted by CLIP~\cite{clip}.
    \item \textbf{User Study}. We invite 31 volunteers and randomly select 40 images from Extended COCO-Stuff dataset to conduct user studies on both automatic and language-based methods. The users are required to choose at most 5 images in each question according to realism, color distortions (color overflow, color artifacts, \textit{etc.}), and text consistency for language-based methods.
\end{itemize}

We present PSNR, SSIM or LPIPS metrics in our Supplementary Materials and further argue that they are not suitable for colorization due to color ambiguity. 

\vspace{-3mm}
\subsubsection{Hue-invariant FID} 
\label{fid_correct}
As many diffusion-based methods~\cite{lcad, coco-lc, CtrlColor, Diffusing-Colors} could generate much more colorful images than ground truth dataset, original FID may be affected by hue difference but not specific color distortions.
To mitigate this problem, we propose a novel \textbf{hue-invariant FID} metric, noted by HI-FID, which calculates FID between ground truth images $I_0$ and color corrected colorized images $\hat{I}_1$. 
Specifically, we adopt the color augmentation method in~\cite{bigcolor}, which scales chromaticity of colorized images $I_1$ in YUV color space as $\mathcal{F}_\alpha: \{\alpha U, \alpha V\} \rightarrow \{U,V\}$, where $\alpha$ denotes the scale factor of saturation. 
We want to minimize the difference of colorfulness by adjusting $\alpha$, so that FID can focus more on local color mistakes rather the overall color tones discrepancy. The whole process can be formulated by
\begin{equation}
    \label{color_corrected}
    \begin{aligned}
    \alpha^* &= \arg \min_\alpha {\Large |} CF(\mathcal{F}_\alpha(I_1){\large )} - CF(I_0){\large |}, \\
    \hat{I}_1 &= \mathcal{F}_{\alpha^*}(I_1),
    \end{aligned}
\end{equation}
where $CF(\cdot)$ denotes the colorfulness metric.

\vspace{-3mm}
\subsubsection{Discussion}
\label{colorfulness_error_text}

\begin{figure}[h]
    \centering
    \includegraphics[width=0.8\linewidth]{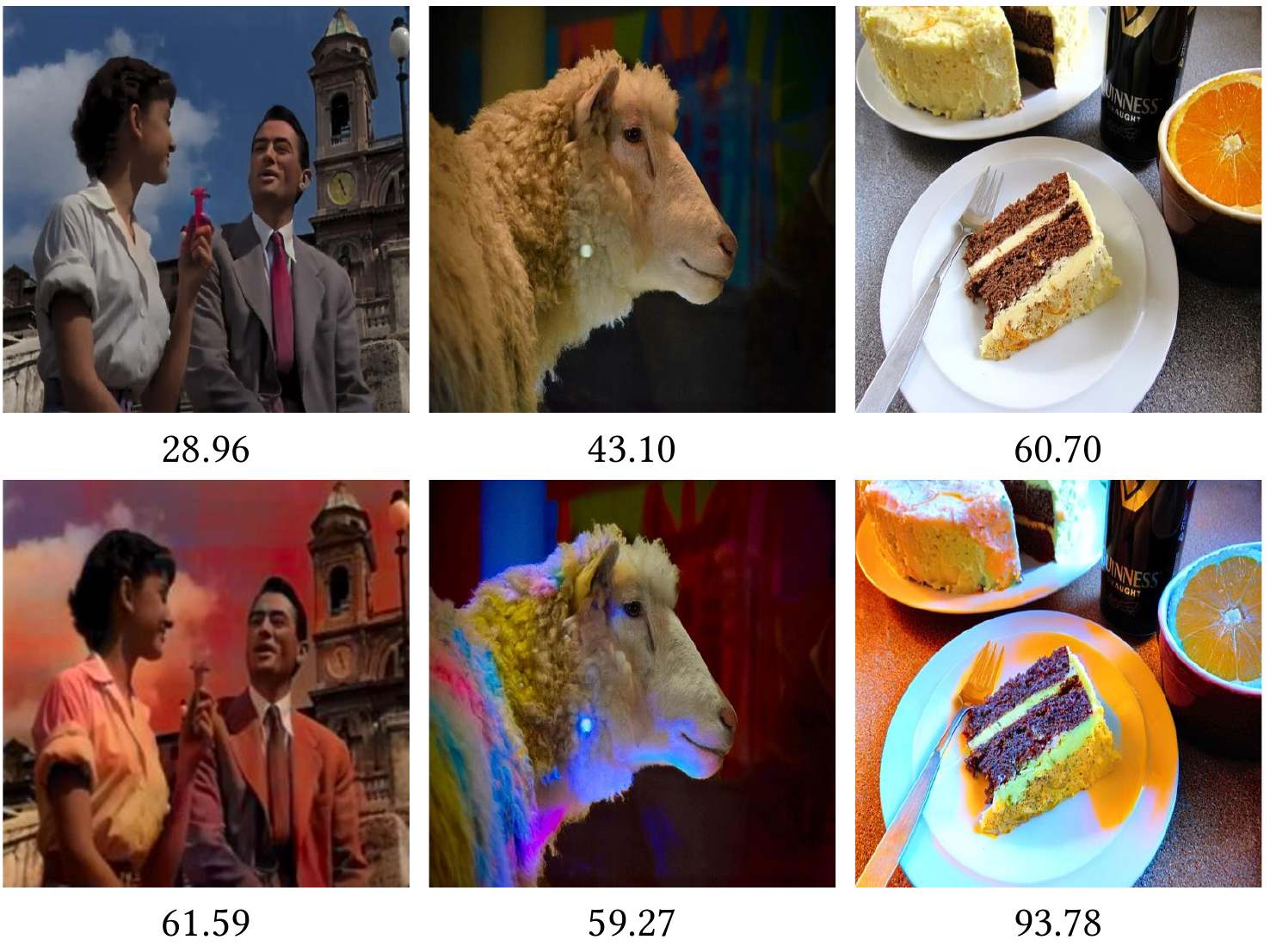}\vspace{-4mm}
    \caption{Failure cases of colorfulness metric. The number under each image is colorfulness score. Higher colorfulness scores are achieved by unpleasant color artifacts such as blue ghosting or unrealistic colors.}
    \vspace{-3mm}
    \label{colorfulness_error}
\end{figure}

Although the colorfulness metric is close to human vision perception, Fig.~\ref{colorfulness_error} shows some failure cases. Some unreal color artifacts like the red sky, blue ghosting and over-saturation can impact the colorfulness score significantly. 

\begin{figure}
\vspace{-5mm}
    \centering
    \includegraphics[width=\linewidth]{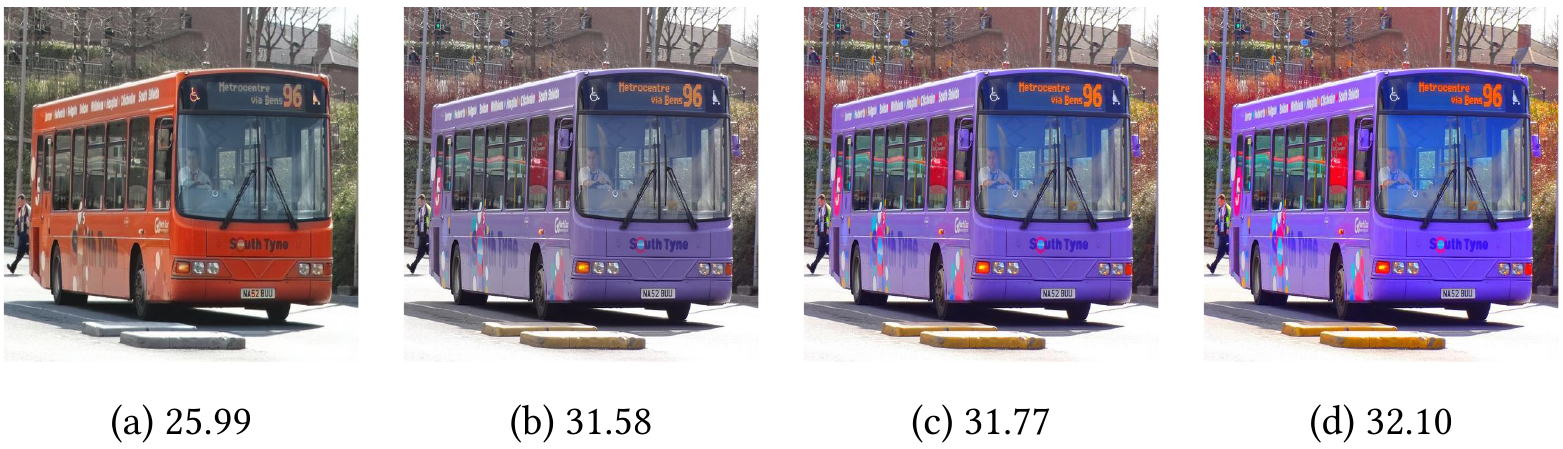}\vspace{-4mm}
    \caption{Visualization of CLIP score. The number under each image denotes its CLIP score. Figure~(a) is produced by the proposed Color-Turbo by prompt ``\textit{a big \textcolor{orange}{orange} bus parking in the plot}". We calculate CLIP score use the prompt ``\textit{a big \textcolor{violet}{purple} bus parking in the plot}".}
    \vspace{-4mm}
    \label{clip_score}
\end{figure}

Besides, CLIP score can only focus on the alignment of high-level semantic information, except for saturation or color distortion.
Although CLIP can align image and text properly, it is still affected by unnatural and over-saturated images, leading to inaccurate measurements. In Fig.~\ref{clip_score}, 
it gives the highest score to the over-saturated image~(d), but lower scores to more natural images~(b) and (c), indicating that CLIP is weak to measure the authenticity of the images. 

\vspace{-3mm}
\subsection{Implementation Details}
We initialize our model with Stable-Diffusion v2.1. 
We experimentally find that our model is robust to different generative backbones, as shown at Sec.~\ref{ablation_backbone}.
We simply set $\lambda_{pixel}=1, \lambda_{lpips}=10, \lambda_{adv}=0.5, \lambda_{clip}=5$ without specific hyperparameter tuning.

\label{cnet}
To complete the discussion of baselines and more clearly benchmark the superior performance of diffusion-based colorization methods blooming recently, we conduct the training of a simple ControlNet, which takes gray images as the input of conditional branch and use color images as ground truth, named \textbf{ControlNet-color}.

Considering that previous methods will produce different resolution,
we further adopt \textbf{luminance replacement strategy}. We resize colorized images to $512\times512$ and convert them to \textit{CIELAB} color space. We then replace their luminance channel with the one of grayscale inputs to maintain the brightness consistency.

\vspace{-4mm}
\subsection{Benchmark on Automatic Image Colorization}
To make fair comparison of both automatic colorization methods and language-based colorization methods in a unified perspective, we wrap all language-based colorization methods with an image captioner, BLIP~\cite{blip}, to provide proper and essential text prompts as the additional input.

\begin{figure}
    \centering
    \includegraphics[width=\linewidth]{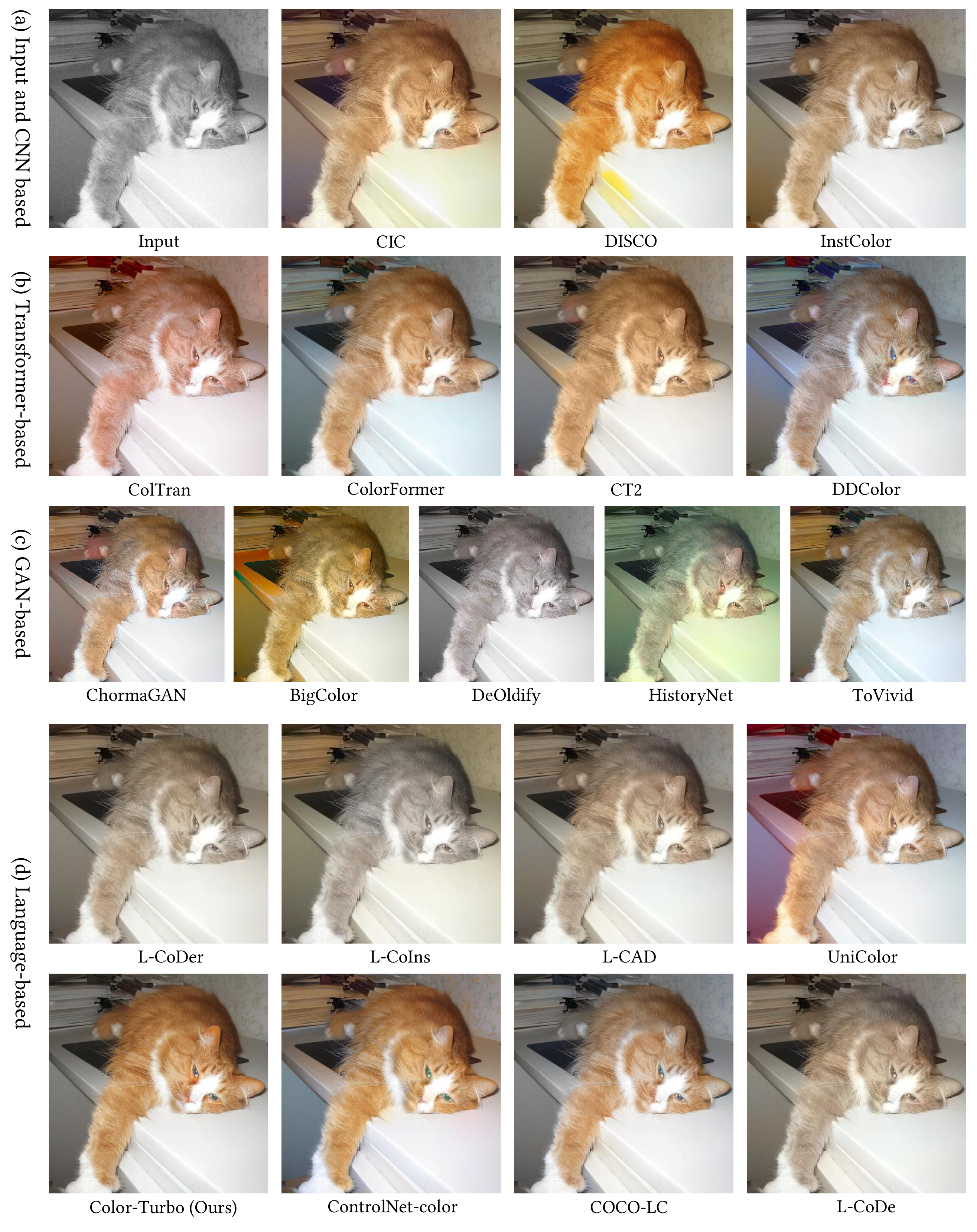}
    \caption{Qualitative results of automatic and language-based methods on automatic colorization task. The text prompt generated by BLIP is "\textit{there is a cat that is laying on top of a desk}".}
    \vspace{-6mm}
    \label{fig:qualitative_uncond}
\end{figure}
We show some representative results in Fig.~\ref{fig:qualitative_uncond}.
In Fig.~\ref{fig:qualitative_uncond} (a), CNN-based methods~\cite{cic, instColor} suffer from under-saturated results. Since DISCO~\cite{DISCO} conducts a two-stage colorization framework where anchor points are firstly colorized, the improper color or number of anchor points lead to color overflow. 
In Fig.~\ref{fig:qualitative_uncond} (b), Transformer-based methods~\cite{colorformer, ct2, ddcolor} struggle to generate high-saturated results. 
Moreover, DDColor~\cite{ddcolor} produces some weird blue artifacts.
In Fig.~\ref{fig:qualitative_uncond} (c), although DeOldfy~\cite{DeOldify} and HistoryNet~\cite{HistoryNet} empower adversarial training, they produce severe grayish results, while BigColor~\cite{bigcolor} suffers from color overflow due to unstable training.
Fig.~\ref{fig:qualitative_uncond} (d) presents, since there are no color words in the text prompt, some color word-based methods~\cite{lcode, lcoder, lcoins} cannot generate colorful results.
COCO-LC, ControlNet-color baseline and our Color-Turbo show their superior performance to extract rich semantic information from grayscale images and simple text prompts without specific color words, while other methods~\cite{lcad, tovivid, unicolor} produce lower-saturated results.

We show more quantitative results in Table~\ref{quan-uncond}. DISCO~\cite{DISCO} and BigColor~\cite{bigcolor} achieve better colorfulness score than other CNN-based or GAN-based methods, which may however come from severe color overflow as analyzed at Sec.~\ref{colorfulness_error_text}.
Although diffusion-based cannot achieve lower FID as Transformer-based method because of different color hues, they get higher ranks under our hue-invariant FID. 

\begin{table*}[t]
\caption{Quantitative results on automatic colorization task. (HI-FID: Hue-invariant FID. CF: Colorfulness. $\Delta$CF: $\Delta$ Colorfulness.)}
\vspace{-3mm}
\label{quan-uncond}
\resizebox{\linewidth}{!}{
\begin{tabular}{c|c|c|cccc|cccc|cccc}
\toprule
\multirow{2}{*}{Methods} & \multirow{2}{*}{category} & \multirow{2}{*}{\thead{User\\ Study}} & \multicolumn{4}{c|}{Extended COCO-Stuff   Dataset} & \multicolumn{4}{c|}{Multi-instances Dataset} & \multicolumn{4}{c}{ImageNet-val5k} \\
 &  &  & FID$\downarrow$ & HI-FID$\downarrow$ & CF $\uparrow$ & $\Delta$CF$\downarrow$ & FID$\downarrow$ & HI-FID $\downarrow$ & CF$\uparrow$ & $\Delta$CF$\downarrow$ & FID$\downarrow$ & HI-FID$\downarrow$ & CF$\uparrow$ & $\Delta$CF$\downarrow$ \\ \midrule
\multicolumn{15}{c}{\textit{Automatic methods}} \\
CIC~\cite{cic} & \multirow{3}{*}{CNN-based} & 0.99 & 29.05 & 28.74 & 29.86 & 13.6 & 17.55 & 16.93 & 26.66 & 13.67 & 13.35 & 12.89 & 25.95 & 11.73 \\
 InstColor~\cite{instColor} &  & 0.66 & 22.82 & 20.21 & 29.94 & 13.52 & 18.2 & 17.66 & 28.5 & 11.83 & 14.39 & 13.29 & 25.96 & 11.72 \\
 DISCO~\cite{DISCO} &  & 3.63 & 23.9 & 19.02 & 51.7 & 8.24 & 14.62 & 11.57 & 48.2s1 & 7.88 & 11.08 & 8.23 & 47.86 & 10.18 \\ \cmidrule{1-15} 
 ColTran~\cite{coltran} & \multirow{4}{*}{Transformer-based} & 1.98 & 35.23 & 32.97 & 31.47 & 11.99 & 25.78 & 22.44 & 33.95 & 6.38 & 20.07 & 18.96 & 28.13 & 9.55 \\
 CT2~\cite{ct2} &  & 2.31 & 18.21 & 16.27 & 25.31 & 18.15 & 10.64 & 13.88 & 41.02 & \underline{0.69} & 7.51 & 10.63 & 39.56 & \underline{1.88} \\
 ColorFormer~\cite{colorformer} &  & 2.97 & 16.84 & 15.73 & 41.52 & \textbf{1.94} & 6.21 & 8.94 & 38.25 & 2.08 & \underline{6.22} & 7.99 & 38.91 & \textbf{1.23} \\
 DDColor~\cite{ddcolor} &  & 8.58 & 14.94 & 14.72 & 40.85 & 2.61 & \textbf{4.89} & \underline{5.26} & 43.5 & 3.17 & \textbf{5.56} & 6.72 & 42.35 & 4.67 \\ \cmidrule{1-15}
 Deoldify~\cite{DeOldify} & \multirow{4}{*}{GAN-based} & 0.99 & 21.21 & 23.76 & 24.51 & 18.95 & 13.74 & 14.03 & 24.75 & 15.58 & 11.02 & 10.55 & 23.89 & 13.79 \\
 HistoryNet~\cite{HistoryNet} &  & 0.66 & 44.26 & 40.28 & 18.01 & 25.45 & 33.2 & 29.89 & 17.51 & 22.82 & 22.38 & 20.35 & 18.02 & 19.66 \\
 ToVivid~\cite{tovivid} &  & 2.97 & 18.33 & 16.29 & 37.53 & 5.93 & 6.9 & 7.11 & 35.46 & 13.35 & 6.69 & 7.01 & 34.67 & 3.01 \\
 ChormaGAN~\cite{ChromaGAN} &  & 2.64 & 35.72 & 32.93 & 28.89 & 14.57 & 25.05 & 24.72 & 26.98 & 13.35 & 16.01 & 18.92 & 26.76 & 10.92 \\
 BigColor~\cite{bigcolor} &  & 6.27 & 27.21 & 25.11 & 48.32 & 4.86 & 11.85 & 10.98 & 47.36 & 7.03 & 12.18 & 10.67 & 44.57 & 6.89 \\ \midrule
\multicolumn{15}{c}{\textit{Language-based methods}} \\
 L-CoDe~\cite{lcode} & \multirow{3}{*}{\thead{from scratch\\ cross-modality}} & 1.32 & 33.33 & 32.79 & 25.9 & 17.56 & 23.77 & 23.02 & 23.57 & 16.76 & 17.96 & 16.54 & 21.37 & 16.31 \\
 L-CoDer~\cite{lcoder} &  & 1.98 & 31.21 & 30.02 & 29.48 & 13.98 & 24.17 & 23.61 & 24.69 & 15.64 & 18.25 & 17.53 & 20.85 & 16.83 \\
 L-CoIns~\cite{lcoins} &  & 2.64 & 33.67 & 32.25 & 28.22 & 15.24 & 24.62 & 22.9 & 22.57 & 17.76 & 17.05 & 15.94 & 21.61 & 16.07 \\ \cmidrule{1-15} 
 UniColor~\cite{unicolor} & \multirow{6}{*}{\thead{pretrained \\ cross-modality}} & 5.94 & 18.82 & 16.67 & 35.61 & 7.85 & 22.4 & 21.23 & 34.28 & 6.05 & 9.7 & 8.33 & 34.28 & 3.4 \\
 L-CAD~\cite{lcad} &  & \underline{11.88} & \textbf{10.74} & \textbf{7.94} & 38.62 & 4.84 & \underline{5.32} & \textbf{4.97} & 32.8 & 7.53 & 7.83 & 6.22 & 29.74 & 7.94 \\
 COCO-LC~\cite{coco-lc} &  & 9.24 & 16.75 & \underline{11.46} & \textbf{58.97} & 15.51 & 11.22 & 7.02 & \textbf{53.12} & 12.79 & 9.63 & 6.58 & \textbf{52.1} & 14.42 \\
 ControlNet-color &  & 7.26 & 15.41 & 12.33 & \underline{51.79} & 7.83 & 12.95 & 9.77 & \underline{49.54} & 9.21 & 7.88 & 6.13 & \underline{48.69} & 11.01 \\
 Color-Turbo ($\alpha=0.5$) &  & 10.23 & 14.92 & 14.2 & 40.87 & \underline{2.59} & 11.63 & 11.04 & 39.97 & \textbf{0.36} & 7.96 & \underline{5.95} & 41.26 & 3.58 \\
 Color-Turbo ($\alpha=1$) &  & \textbf{14.85} & \underline{14.03} & 13.92 & 33.87 & 9.59 & 11.01 & 10.86 & 34.53 & 5.8 & 6.28 & \textbf{4.39} & 35.07 & 2.61 \\

\bottomrule
\end{tabular}
}
\vspace{-4mm}
\end{table*}

\vspace{-3mm}
\subsection{Benchmark on Language-based Image Colorization}
\subsubsection{Qualitative results.}
\begin{figure*}[h]
    \centering
    \includegraphics[width=\linewidth]{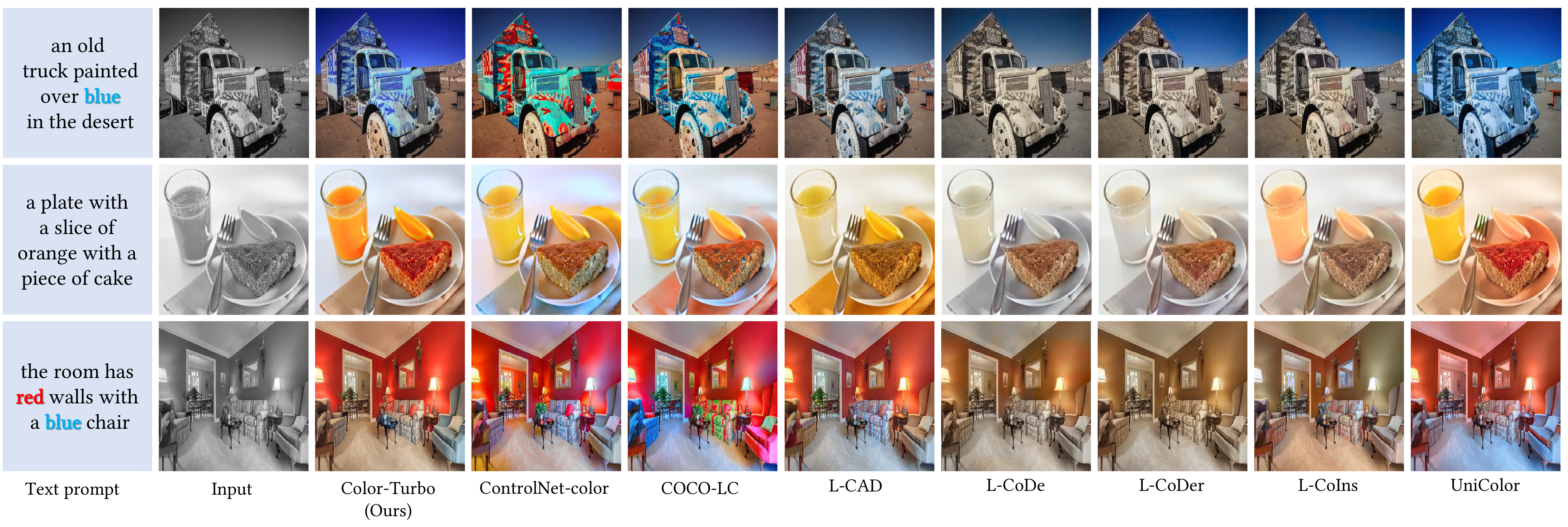}\vspace{-4mm}
    \caption{Qualitative results of language-based methods. Zoom in for better visualization.}
    \vspace{-4mm}
    \label{fig:qualitative_cond}
\end{figure*}
In Fig.~\ref{fig:qualitative_cond}, we use (1) ``blue old truck'', (2) ``orange and cake'', and (3) ``red walls blue chair'',
to illustrate the performance of the language-based methods.
Methods that train a cross-modality model from scratch~\cite{lcode, lcoder, lcoins} are hard to produce plausible images, and struggle to understand text prompts precisely, leading to unrealistic and incorrect colorized results.
Although ControlNet-color can generate high-saturated colorized images, it produces non-negligible color artifacts and color overflow, especially noitceable as blue ghosting under white background. Besides, ControlNet-color sometimes "over-creates" misaligned colors with luminance conditions, as shown in the first example in Fig.~\ref{fig:qualitative_cond}, where some weird color blocks occur on the background. Despite ControlNet-color cannot achieve perfect performance, it demonstrates the superior of diffusion models on language-based colorization.

Since COCO-LC~\cite{coco-lc} adopts the ControlNet-based architecture, it also generates color artifacts as shown in the third example, but it alleviates such drawback compared with ControlNet-color. 
L-CAD~\cite{lcad} finetunes Stable Diffusion and inserts grayscale conditions into VAE decoder, leading to lower saturated results, but effectively reducing color artifacts. Unluckily, L-CAD also produces slight color overflow such as the bleeding yellow around the orange in the second example.
UniColor~\cite{unicolor} can produce colorful images with the benefits of hybrid-Transformer and Chroma-VQGAN with adversarial training, but it suffers from color overflow because the CLIP-based alignment is not spatially precise, as shown as the blue wheel, brown plate and red ceil in Fig.~\ref{fig:qualitative_cond} from top to bottom.
Compared with all previous methods and ControlNet-color baseline, our Color-Turbo generates the most realistic results with minimal color distortions, and match the requirements of text prompt precisely.

\vspace{-3mm}
\subsubsection{Quantitative results.}
Table~\ref{quan-cond} presents quantitative results of UniColor~\cite{unicolor}, L-CoDe~\cite{lcode}, L-CoDer~\cite{lcoder}, L-CoIns~\cite{lcoins}, L-CAD~\cite{lcad}, COCO-LC~\cite{coco-lc}, basic ControlNet baseline for image colorization mentioned at Sec.~\ref{cnet} and our proposed Color-Turbo. 
Color-Turbo gets the best CLIP score among three evaluation dataset, and comparable FID, hue-invariant FID score, demonstrating its promising performance.
Although Color-Turbo cannot achieve high colorfulness score as other diffusion-based methods, those methods with high colorfulness may suffer from color overflow and color artifacts as analyzed at Sec.~\ref{colorfulness_error_text} and Fig.~\ref{colorfulness_error}.
Moreover, Color-Turbo is significantly more efficient and can produce one image in about 0.33s, more than 14 times faster than the other diffusion-based methods. We show more inference time comparison in our Supplementary Material.

\begin{table*}[h]
\caption{Quantitative results of language-based colorization methods. (HI-FID: Hue-invariant FID. CF: Colorfulness. $\Delta$CF: $\Delta$ Colorfulness.)}
\vspace{-3mm}
\label{quan-cond}
\resizebox{\linewidth}{!}{
\begin{tabular}{c|c|ccccc|ccccc|ccccc}
\toprule
\multirow{2}{*}{Methods} & \multirow{2}{*}{\thead{User \\ Study}} & \multicolumn{5}{c|}{Extended COCO-Stuff Dataset} & \multicolumn{5}{c|}{Multi-instances Dataset} & \multicolumn{5}{c}{ImageNet-val5k} \\  
 & & FID$\downarrow$ & HI-FID$\downarrow$ & CF$\uparrow$ & $\Delta$CF$\downarrow$ & \multicolumn{1}{c|}{CLIP Score$\uparrow$} & FID$\downarrow$ & HI-FID$\downarrow$ & CF$\uparrow$ & $\Delta$CF$\downarrow$ & CLIP Score$\uparrow$ & FID$\downarrow$ & HI-FID$\downarrow$ & CF$\uparrow$ & $\Delta$CF$\downarrow$ & CLIP Score$\uparrow$ \\ \midrule
 \multicolumn{17}{c}{\textit{From scratch cross-modality methods}} \\
L-CoDe~\cite{lcode} & 2.78 & 31.21 & 29.92 & 29.48 & 13.97 & \multicolumn{1}{c|}{29.78} & 25.08 & 22.73 & 25.18 & 16.13 & 27.59 & 19.56 & 15.15 & 24.23 & 12.41 & 27.66 \\
L-CoDer~\cite{lcoder} & 3.89 & 31.78 & 29.98 & 29.52 & 13.93 & \multicolumn{1}{c|}{29.98} & 22.75 & 19.09 & 26.45 & 14.86 & 27.53 & 16.73 & 13.78 & 22.9 & 13.74 & 27.93 \\
L-CoIns~\cite{lcoins}& 7.22 & 33.67 & 30.27 & 25.31 & 18.14 & \multicolumn{1}{c|}{30.06} & 23.13 & 21.64 & 30.65 & 10.66 & 27.61 & 21.03 & 14.24 & 33.53 & 3.11 & 28.57 \\ \cmidrule{1-17} \multicolumn{17}{c}{\textit{Pretrained cross-modality methods}} \\
UniColor~\cite{unicolor} & 9.44 & 16.03 & 13.41 & 41.40 & \underline{2.05} & \multicolumn{1}{c|}{26.43} & 19.32 & 16.78 & 33.53 & 7.78 & 26.42 & 10.63 & 7.83 & 34.98 & 1.66 & 28.40 \\
L-CAD~\cite{lcad} & \underline{16.67} & \textbf{12.75} & \textbf{13.04} & 43.80 & \textbf{0.35} & \multicolumn{1}{c|}{30.7} & 7.47 & 7.89 & 33.27 & 8.04 & 29.34 & 9.30 & 9.42 & 36.34 & \textbf{0.30} & 29.02 \\
COCO-LC~\cite{coco-lc} & 15 & 16.5 & 15.67 & \textbf{54.71} & 11.26 & \multicolumn{1}{c|}{30.88} & 9.77 & 7.98 & \underline{48.46} & \underline{7.15} & 28.66 & 9.86 & 8.92 & \underline{43.01} & 6.37 & 29.32 \\
ControlNet-color & 16.13 & 16.45 & 16.73 & \underline{53.33} & 9.88 & \multicolumn{1}{c|}{30.54} & 11.62 & 11.21 & \textbf{50.49} & 9.18 & 28.49 & 8.93 & 8.27 & \textbf{46.73} & 10.09 & 29.22 \\
Color-Turbo ($\alpha=0.5$) & 13.89 & 13.89 & \underline{13.26} & 39.45 & 4.00 & \multicolumn{1}{c|}{\underline{33.02}} & \underline{7.25} & \underline{6.73} & 36.38 & \textbf{4.93} & \underline{31.29} & \underline{6.34}  & \textbf{6.13} & 35.93 & \underline{0.71} & \textbf{31.20} \\
Color-Turbo ($\alpha=1$) & \textbf{17.78} & \underline{13.63} & 13.31 & 33.35 & 10.10 & \multicolumn{1}{c|}{\textbf{33.23}} & \textbf{5.82} & \textbf{6.25} & 26.29 & 15.02 & \textbf{31.05} & \textbf{6.28} & \underline{6.43} & 27.19 & 6.45 & \underline{30.72} \\

\bottomrule
\end{tabular}
}
\end{table*}

\vspace{-4mm}
\subsection{User Study}
Table~\ref{quan-uncond} and Table~\ref{quan-cond} present the results of user study on automatic and language-based colorization task. 
Diffusion-based methods~\cite{lcad, coco-lc}, ControlNet-color and Color-Turbo perform better than other methods on both automatic and language-based colorization task.
Notably, HI-FID matches more accurately with users' preference than FID on diffusion-based methods, providing a fair comparison of the methods which generate high-saturated results.

\vspace{-3mm}
\subsection{Performance Analysis of Color-Turbo}
In addition to method benchmarking, we conduct experiments to further analyze the performance of Color-Turbo.

\vspace{-3mm}
\subsubsection{Ablation Study}
\begin{figure}[h]
    \centering
    \includegraphics[width=\linewidth]{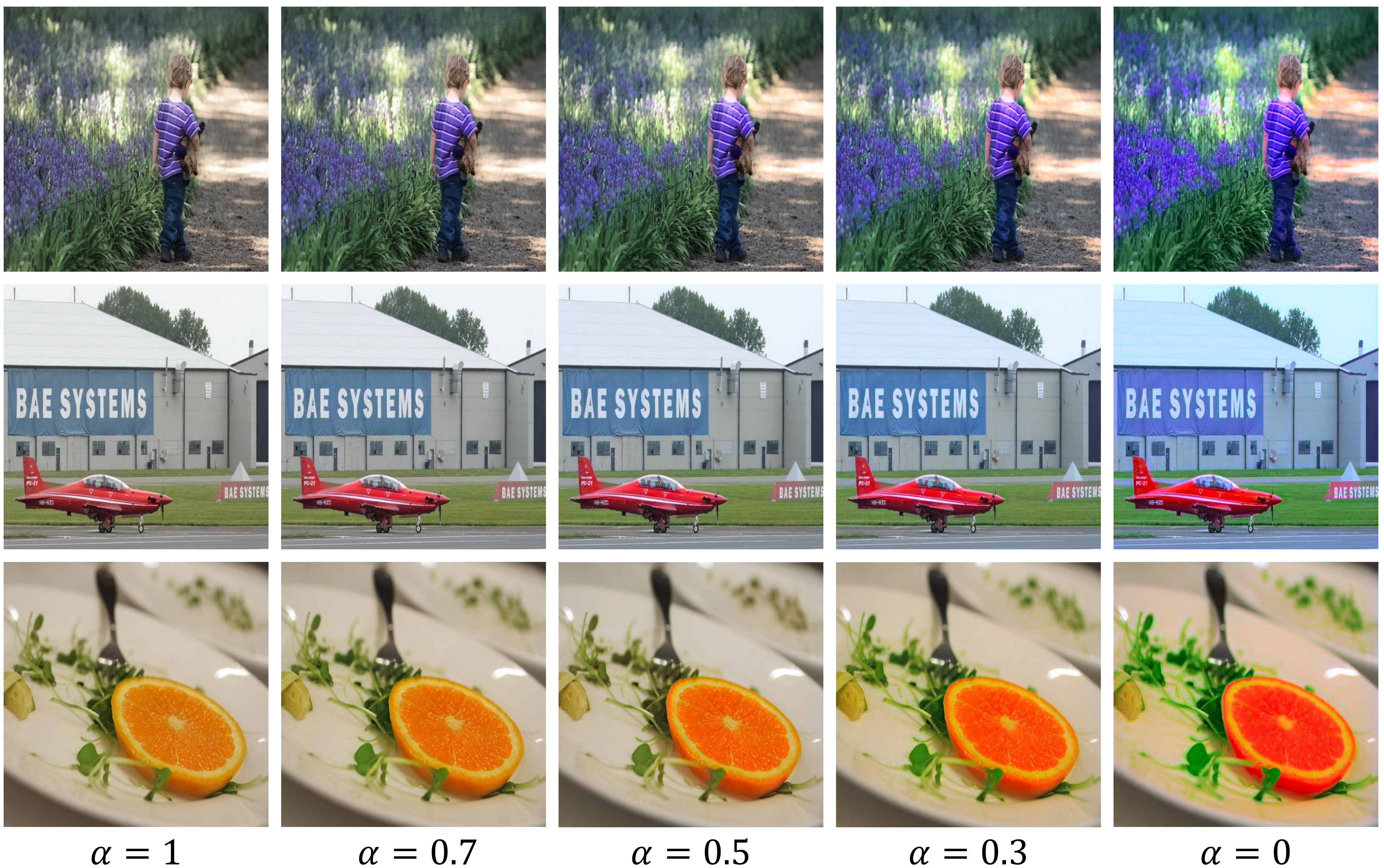}\vspace{-3mm}
    \caption{Diverse results with different scale factor. Lower alpha, less grayscale intensity, leads to higher saturation.}
    \vspace{-4mm}
    \label{fig:ablation_alpha}
\end{figure}

\textbf{Scale factors.}
We experiment on different scale factors in Fig.~\ref{fig:ablation_alpha}. The smaller scale factor is, the less grayscale information is injected into the decoder, increasing the colorfulness. However, as less grayscale conditions are considered during decoding, the structure of colorized images cannot be maintained well, leading to color overflow. 

\begin{figure}[h]
    \centering
    \includegraphics[width=0.9\linewidth]{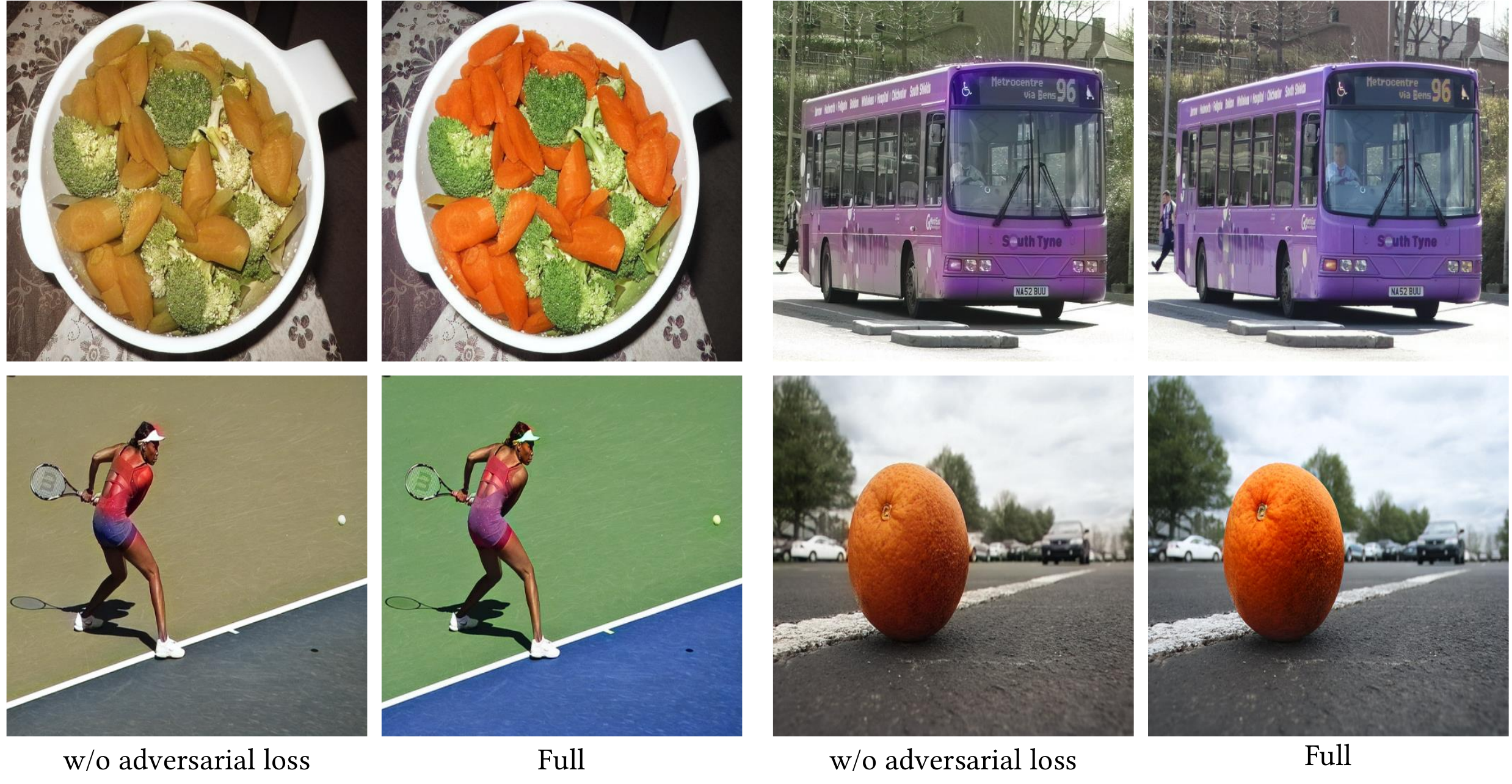}\vspace{-3mm}
    \caption{Ablation results of adversarial loss. The results are more vivid and colorful with adversarial loss.}
    \vspace{-4mm}
    \label{fig:ablation_adv}
\end{figure}

\textbf{Adversarial loss.} 
We study the effect of adversarial training in Fig.~\ref{fig:ablation_adv}. The results will be grayish and under-saturated without adversarial loss, because simple regression losses encourage conservative results such as the mean of the colors, as analyzed at Sec.~\ref{cic}.

\textbf{Generative backbone.} Table~\ref{Table:ablation_backbone}
\label{ablation_backbone}
presents quantitative results with few differences between SDv1.5 and SDv2.1, indicating that our Color-Turbo is robust to different backbones.
\vspace{-5mm}
\begin{table}[h]
\centering
\caption{Ablation results of different generative backbone.}
\vspace{-3mm}
\label{Table:ablation_backbone}
\resizebox{\linewidth}{!}{
\begin{tabular}{c|ccc|ccc|ccc}
\toprule
\multirow{2}{*}{Backbone} & \multicolumn{3}{c|}{Extended COCO-Stuff Dataset} & \multicolumn{3}{c|}{Multi-instances Dataset} & \multicolumn{3}{c}{ImageNet-val5k} \\  
 & FID$\downarrow$ & CF$\uparrow$ & \multicolumn{1}{c|}{CLIP Score$\uparrow$} & FID$\downarrow$ & CF$\uparrow$ & CLIP Score$\uparrow$ & FID$\downarrow$ & CF$\uparrow$ & CLIP Score$\uparrow$ \\ \midrule
SDv1.5 & 13.63 & 31.69 & 33.28 & 5.84 & 26.43 & 31.13 & 6.30 & 27.33 & 30.77 \\
SDv2.1 (Ours) & 13.63 & 33.35 & 33.23 & 5.82 & 26.29 & 31.05 & 6.28 & 27.19 & 30.72 \\
\bottomrule
\end{tabular}
}
\vspace{-5mm}
\end{table}

\vspace{-2mm}
\subsubsection{Limitations and Discussions}
Although Color-Turbo can achieve efficient, accurate and controllable language-based colorization, there are several crucial limitations to address in the future works.

\noindent \textbf{Diverse results.} Color-Turbo cannot generate diverse colorized images. Initial random noise is a double-edge sword, which can generate results with color overflows and artifacts as shown in Fig.~\ref{diff_seed}, but can also generate creative results to cater different users' appetites. 

\begin{figure}[h]
    \centering
    \includegraphics[width=\linewidth]{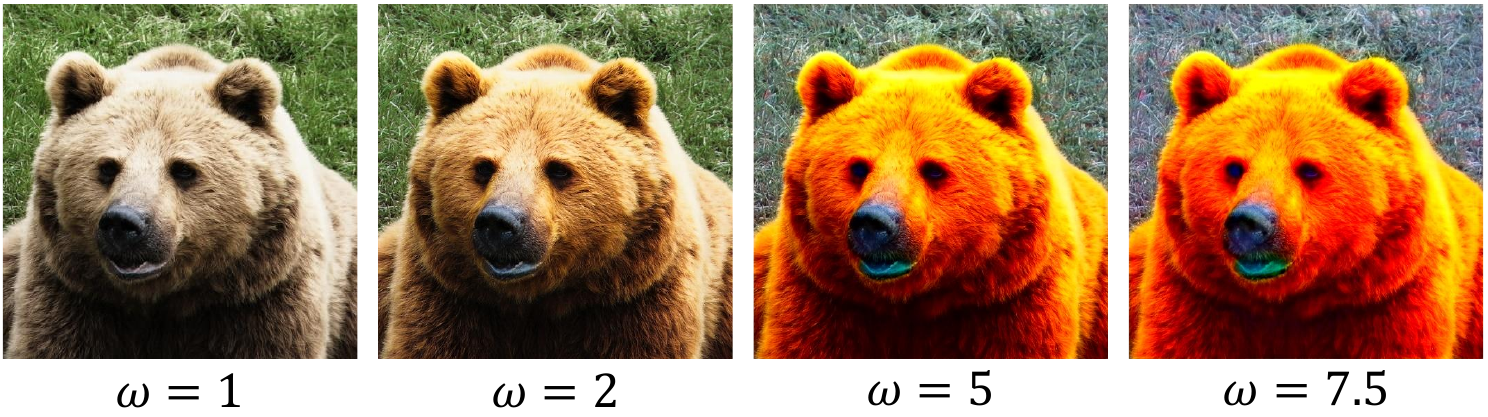}
    \vspace{-6mm}
    \caption{CFG results of our Color-Turbo with different intensity $w$. Specifically, $w=1$ represents no use of CFG, and $w=7.5$ means common practice of Stable Diffusion model. Prompt: ``the large brown bear has a black nose". Negative prompt: ``grayish".}
    \vspace{1mm}
    \label{color-turbo-cfg}
    \centering
    \includegraphics[width=0.9\linewidth]{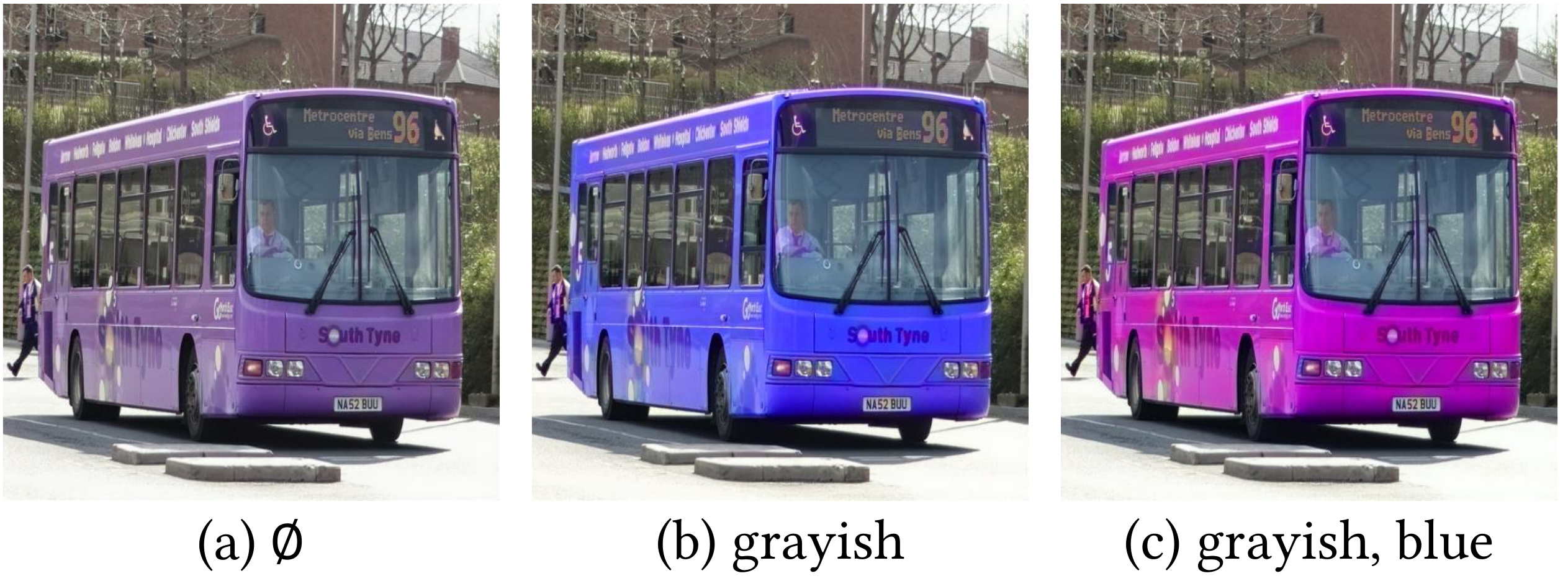}\vspace{-3mm}
    \caption{CFG of Color-Turbo sometimes causes color error. Prompt: ``a big purple bus parking in the plot''. The captions represent the negative prompts.}
    \vspace{1mm}
    \label{color-turbo-cfg-2}
    \centering
    \includegraphics[width=\linewidth]{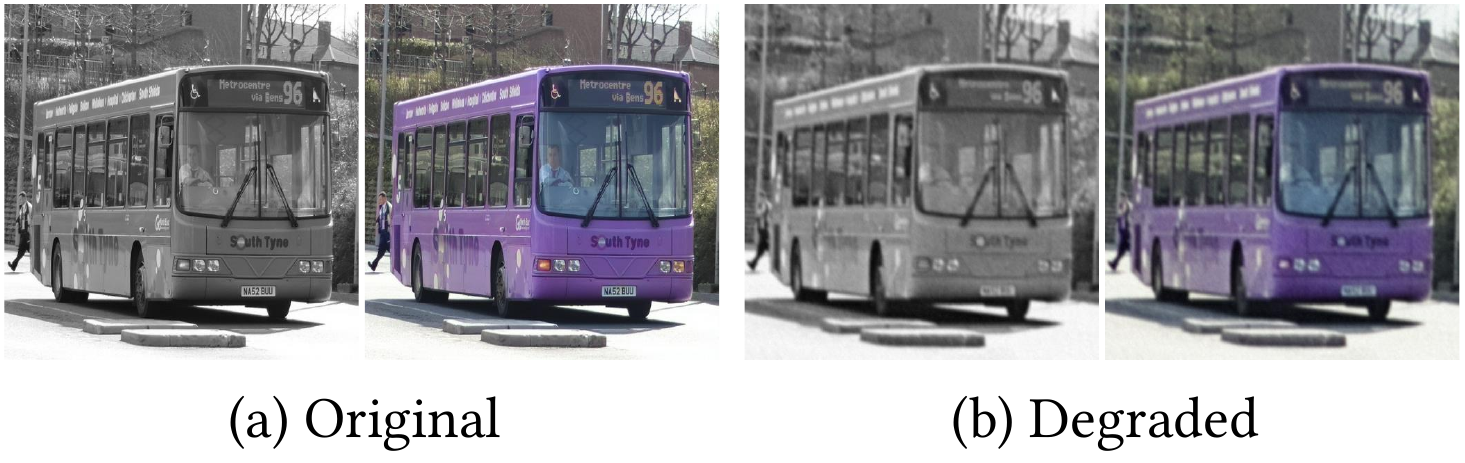}\vspace{-5mm}
    \caption{(a) Results of Color-Turbo under normal conditions withour degradations. (b) Color-Turbo can neither perform well under degraded conditions as the face of bus driver becomes grayish, nor restore a clear image from the degraded one.}
    \vspace{-7mm}
    \label{degrade}
\end{figure}

\noindent \textbf{Classifier-free guidance.} \ys{Color-Turbo cannot directly reuse the setting of classifier-free guidance~\cite{cfg} (CFG) as in the original Stable Diffusion.  
We provide an example in Fig.~\ref{color-turbo-cfg} with different CFG intensity $w$. We experimentally find that our best CFG intensity is lower than $w=7.5$ in the common Stable Diffusion setting.}
Moreover, there might be color errors when Color-Turbo utilizes CFG as shown in Fig.~\ref{color-turbo-cfg-2}. Simple negative prompts such as ``grayish" will change the purple bus to blue, while adding ``blue" back to negative prompts maintains the proper color. Utilizing CFG in Color-Turbo requires more modifications. 
\vspace{-3mm}
\vspace{-2mm}
\section{Conclusions and Future Directions}
\label{sec:conclusion}
This paper presents a comprehensive survey on language-based image colorization methods and a rigorous benchmark on automatic and language-based colorization tasks.
We briefly review existing automatic methods following the model architectures and training paradigms, and uncover their crucial problems including color ambiguity and insufficient controllability.
As for language-based methods, we argue that the alignment of language descriptions and image regions are crucial points, and thoroughly summarize different condition insertion methods under the formulation of conditional generation. We reveal the difficulty to achieve an optimal balance between color generation and structural consistency.
In view of above limitations, we propose a new baseline for efficient and effective language-based colorization, named Color-Turbo. Revisiting the strong generative ability of diffusion models, our method can produce plausible and stable colorized images with the least inference cost.
Extensive experiments on benchmark datasets demonstrate our superior performance and generalization capacity.
We propose a novel and effective metric based on FID, the hue-invariant FID,  which can more accurately measure the distortion of local color structure without the bias of global hue deviation. Further user study demonstrate its effectiveness.

For the future research on language-based colorization, the following pending issues deserve more attention:
\begin{itemize}
    \item \textbf{Stable and diverse colorization.} 
    Under the formulation of diffusion-based controllable generation, random noises can lead to diverse results, but probably cause unstable colorization, such as color distortions. On the contrary, taking grayscale images as input will limited to generate diverse colorization results. It is hard to fully exploit potential generative ability of diffusion models while preserve stable generation process.

    \item \textbf{Long inference time.} Although existing diffusion-based methods have demonstrated their superior performance, they are time-consuming because of the inherent diffusion sampling process. Our Color-Turbo mitigates such limitation, but it is still hard to achieve real-time inference due to the large model.

    \item \textbf{Degradation-aware colorization.} 
    Considering that most of legacy images or monochrome films contains inevitable degradations, 
    it is necessary to explore the generalization power of colorization models on degraded environments. 
    As shown in Fig.~\ref{degrade}, Color-Turbo cannot perform well under blur, noises and compression.
    A two stage method, \textit{i.e.} first restoration then colorization is a straightforward solution, but probably not optimal because of the potential accumulative error and increased inference cost of time and memory.
    
\end{itemize}

\vspace{-4mm}
\bibliographystyle{IEEEtran}
\bibliography{ms}

\includepdf[pages=-]{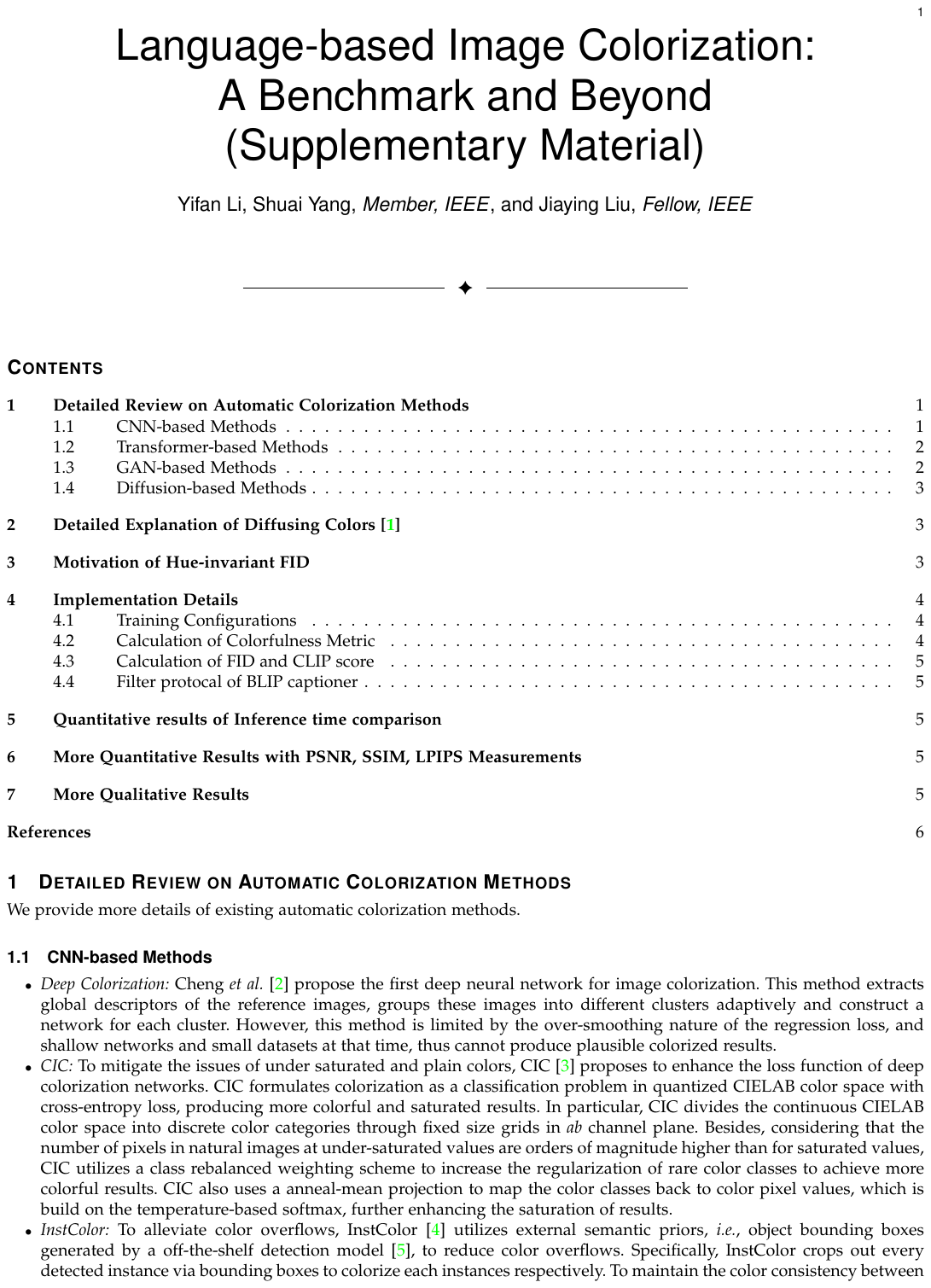}

\end{document}


\title{Language-based Image Colorization:\\ A Benchmark and Beyond \\(Supplementary Material)}

\author{Yifan~Li, 
        Shuai~Yang,~\IEEEmembership{Member,~IEEE},
        and~Jiaying~Liu,~\IEEEmembership{Fellow,~IEEE}
}

\maketitle

\IEEEdisplaynontitleabstractindextext
\IEEEpeerreviewmaketitle
\tableofcontents

\section{Detailed Review on Automatic Colorization Methods}

We provide more details of existing automatic colorization methods. 

\subsection{CNN-based Methods}
\begin{itemize}
    \item \textit{Deep Colorization:} Cheng \etal~\cite{deep_color} propose the first deep neural network for image colorization.
    This method extracts global descriptors of the reference images, groups these images into different clusters adaptively and construct a network for each cluster. 
    However, this method is limited by the over-smoothing nature of the regression loss, and shallow networks and small datasets at that time, thus cannot produce plausible colorized results. 
    \item \textit{CIC:} 
    \label{cic}
    To mitigate the issues of under saturated and plain colors, CIC~\cite{cic} proposes to enhance the loss function of deep colorization networks. CIC formulates colorization as a classification problem in quantized CIELAB color space with cross-entropy loss, producing more colorful and saturated results. In particular, CIC divides the continuous CIELAB color space into discrete color categories through fixed size grids in \textit{ab} channel plane. Besides, considering that the number of pixels in natural images at under-saturated values are orders of magnitude higher than for saturated values,
    CIC utilizes a class rebalanced weighting scheme to increase the regularization of rare color classes to achieve more colorful results. CIC also uses a anneal-mean projection to map the color classes back to color pixel values, which is build on the temperature-based softmax, further enhancing the saturation of results.
    \item \textit{InstColor:} To alleviate color overflows, InstColor~\cite{instColor} utilizes external semantic priors, \textit{i.e.}, object bounding boxes generated by a off-the-shelf detection model~\cite{maskRCNN}, to reduce color overflows. Specifically, InstColor crops out every detected instance via bounding boxes to colorize each instances respectively. To maintain the color consistency between each instances blocks and the overall images, InstColor proposes a fusion module to merge each instances. However, the incorrect external prior extraction may instead aggravate color overflows.
    
    \item \textit{DISCO:} Given the poor performance of single-stage CNN-based colorization methods, DISCO~\cite{DISCO} designs a coarse-to-fine colorization framework. In particular, DISCO use superpixel segmentation~\cite{super_pixel} and K-Means to construct color clusters and sample anchor points. In the coarse stage, a color modeler predicts color distribution for anchor color representation, while in the fine stage, a color generator predicts the remaining pixel colors by referring the sampled anchor colors.
\end{itemize}

\subsection{Transformer-based Methods}
Since Transformer~\cite{transformer} is successfully introduced to vision recognition, Vision Transformer, or ViT~\cite{vit}, has developed rapidly in many downstream vision tasks with its superb ability on long-range modeling.
\begin{itemize}
    \item \textit{Colorization Transformer (ColTran):} ColTran~\cite{coltran} proposes the first transformer-based colorization model based on Axial Transformer and incorporates a multi-stage colorization strategy. ColTran separates the whole colorization process into three consecutive parts. 
    Firstly, a low bit-rate color information and low spatial resolution result is produced by an autoregressive colorizer. 
    Secondly, a color upsampler is utilized to enhance color diversity based on the coarse result. 
    Finally, ColTran uses a spatial upsampler to generate high-resolution images. However, such multi-stage strategy is time-consuming and may accumulate errors. 
    Furthermore, ColTran essentially treats each image pixel as a single token in a pure transformer architechture, lacking of local interaction and inductive bias as CNN.

    \item \textit{ColorFormer:} To reduce computational complexity and make fully usage of long-range modeling ability of Transformer, ColorFormer~\cite{colorformer} proposes a Global-Local hybrid Multi-head Self-Attention operation (GL-MSA) to balance the trade-off between global attention receptive field and computational cost, which concatenates local and global features in \textit{keys} and \textit{values} during the calculation of attention. As for the decoder part, ColorFormer proposes a color memory (CM) module which stored multiple groups of semantic-color mapping for efficient and diverse color acquisition.
    
    \item \textit{CT2:} 
    CT2~\cite{ct2} draws inspirations from CIC~\cite{cic}, and 
    observes that the distribution of $a$, $b$ channels is not full-filled for natural images in CIELAB color space.
    Thus it computes statistics of valid $(a, b)$ pairs on a large-scale natural images dataset and filters out those empirically invalid grids,
    resulting in 313 meaningful color tokens. 
    On the basis of this, CT2 leverages an adaptive attention mechanism to bridge correspondence between such prior color tokens and luminance tokens from input.
    
    \item \textit{DDColor:} Given the strong learning ability of query-based vision transformer~\cite{detr}, DDColor~\cite{ddcolor} presents a color decoder that holds a set of learnable color tokens according to different colors, and an adaptive fusion module to merge the grayscale structure information and the generated color information with the help of cross attention mechanism. By utilizing a pre-trained visual backbone to extract gray images' features, DDColor can better preserve the details in grayscale images. 
    However, DDColor sometimes suffers color confusion due to color ambiguity, \textit{i.e.} a single object is colorized in multiple different but reasonable colors.

    \item \textit{MultiColor:} MultiColor~\cite{MultiColor} builds upon DDColor~\cite{ddcolor} and proposes a multi-branch model to fully use the complementarity from multiple color spaces for image colorization.
    MultiColor facilitates more nuanced color information based on the inconsistent color distributions depicted by different color spaces such as HSV, CIE-LAB and YUV.

\end{itemize}
\subsection{GAN-based Methods}
GANs have achieved remarkable success in synthesizing high-fidelity natural images, leading to its wide application to image restoration tasks, including image colorization. 
\begin{itemize}
    \item \textit{ChromaGAN:} ChromaGAN~\cite{ChromaGAN} designs an encoder-decoder architecture to learn the mapping from grayscale inputs to color images. It simply uses a typical PatchGAN~\cite{patchgan} discriminator with WGAN~\cite{wgan} loss to maintain realism. 
    To achieve precise regularization, ChromaGAN utilizes $L_2$ loss on chrominance channels, \textit{i.e.} $(a,b)$ channels in CIE-LAB color space, to maintain more effective regularization. Besides, ChromaGAN also constructs a semantic class distribution loss based on VGG-16~\cite{vgg} pre-trained on ImageNet~\cite{imagenet}. Specifically, ChromaGAN uses class labels predicted on grayscale images as the ground truth label.
    \item \textit{HistoryNet:} HistoryNet~\cite{HistoryNet} incorporates classification and segmentation into colorization process. In particular, HistoryNet pays special attention to the characteristics of historical images. It presents a classification sub-module to supply information about the eras, nationalities, and garment types. Furthermore, in order to better colorize the persons in historical images, HistoryNet proposes a parsing network and uses human parsings as additional supervisions. It is worth noting that HistoryNet collects a large-scale Modern Historical Movies Dataset (MHMD), which contains 1,353,166 images and the 42 labels of eras, nationalities, and garment types. However, since HistoryNet is trained on data from old movies, it generally produces grayish colorized results.  

    \item \textit{DeOldify:} Deoldify~\cite{DeOldify} proposes an efficient GAN training strategy that adopts an asynchronous training mode on generator and discriminator. Specifically, Deoldify first trains a generator with only a perceptual loss, then trains a discriminator with a classification loss on images generated by the generator, and finally jointly trains generator and discriminator in a GAN setting. Such method can stabilize the training process and eliminate artifacts caused by vanilla GAN. Besides, DeOldify adopts a pretrained U-Net with minor modifications on it to involve self-attention and spectral normalization.

    \item \textit{ToVivid:} ToVivid~\cite{tovivid} proposes to use the generative prior of pre-trained BigGAN~\cite{biggan} in an inversion manner.
    It searches an appropriate latent code of the color image which is aligned with the grayscale input approximately in the BigGAN latent space, then utilizes the GAN-inverted color image generated by such latent code as additional information during colorization. 
    As for spatial alignment between the GAN-inverted color image and the original grayscale input, ToVivid attempts to warp the synthesized color features of GAN-inverted images with spatial correlation matrix between GAN-inverted color images and grayscale input images. 
    However, GAN inversion is time-consuming and inaccurate especially on grayscale images. Besides, the mismatches between GAN-inverted images and input images are inevitable, leading to color distortions.

    \item \textit{BigColor:} In view of above drawbacks, BigColor~\cite{bigcolor} builds an encoder-generator model, which equipped with pre-trained BigGAN's generator and discriminator. 
    It trains an encoder to extract spatial features with an additional classifier to obtain class embeddings, and feeds both of them with rich semantic information of grayscale images into the generator. BigGAN has strong generative ability in general scenes, and could be applied to degraded legacy photos. 
    Unfortunately, it requires a high training cost, and may produce color artifacts.

\end{itemize}

\subsection{Diffusion-based Methods}
\label{sec:uncond_diff}
Diffusion models~\cite{ddpm, sde} convert samples from a standard Gaussian distribution into samples from an empirical data distribution through an iterative denoising process.
The powerful diffusion generative prior is a double-edge sword for image colorization task:
Diffusion models such as Stable Diffusion~\cite{sd} pre-trained on a large-scale dataset can generate high-quality images; However, as diffusion sampling process is stochastic, it is hard to constrain the generated results to be aligned with the grayscale input precisely without sacrificing the original generative ability.

\begin{itemize}
    \item \textit{Palette:} Palette~\cite{saharia2022palette} seizes the powerful fitting ability of diffusion paradigm and trains a diffusion model from scratch. Specifically, Palette adopts a U-Net architecture with convolution blocks to merge grayscale image conditions through concatenation. Although this approach is simple yet effective, it cost a lot to convergence.

    \item \textit{ColorDiff:} To fully utilize the strong generative prior of Stable Diffusion~\cite{sd}, ColorDiff~\cite{colordiff} introduces segmentation maps extracted from grayscale inputs as external priors and designs a high-level semantic insertion module to enhance the model's capability of producing semantically reasonable colors. Besides, to maintain structure consistency, ColorDiff builds skip connections between VAE encoder and decoder to insert grayscale features during the decode process, and adds an additional scale factor to control the intensity of grayscale information insertion flexibly. 
    However, inaccurate external priors will damage the colorization, producing severe color overflows and artifacts. 
    
    \item \textit{Imagine-Colorization:} Cong \etal~\cite{imagination} propose a ControlNet-based colorization framework. The method uses pre-trained ControlNet~\cite{controlnet} to produce several colorized candidate images based on edge maps extracted from grayscale inputs, which are aggregated into a single color reference image with the help of Semantic-SAM~\cite{semantic_sam}.
    Finally the finer colorization phase conducts reference-based colorization by a coarse-to-fine hint points generation and propagation strategy. In the coarse stage, it divides the whole images into several patches and extracts hint points from each of them. 
    In the finer stage, it draws inspirations from UniColor~\cite{unicolor} and adopts a transformer to autoregressively predict chrominance features with luminance features extracted from the inputs.
    This method also enhances controllability by allowing users to select appropriate candidate images. However, multiple candidate images generation is time-consuming. The forward of ControlNet is time-consuming, and such forward need to be done more than once (about 32 times reported in their paper) in this method, obviously increasing total time cost. In addition, the incorrect segmentation results, especially inferred from grayscale images, probably lead to color overflows.
\end{itemize}

\section{Detailed Explanation of Diffusing Colors~\cite{Diffusing-Colors}}
Specifically, Diffusing Colors forms the color enhancement process as a convex combination of gray images 
$x_{gray}$ and color images $x_{color}$, which can be formulated by:
\begin{equation}
    x_t = t \cdot x_{gray} + (1-t) \cdot x_{color}, t \in [0,1],
\end{equation}
where $t$ denotes the timestep of diffusion process, with $t=1$ as the start point of diffusion.

Diffusing Colors inherits the pre-trained U-Net and finetunes it to predict the color residuals that can be used to reconstruct final color images from grayscale inputs. Particularly, the color residuals are defined as $\Delta=x_{color}-x_{gray}$, and at each timestep $\Delta_t=t \cdot \Delta$. 
As for colorization sampling, Diffusing Colors fully adopts the philosophy of DDIM sampling strategy~\cite{ddim}, which predicts an approximate colorized image $\hat{x}_{color}$ by directly utilizing the predicted $\Delta_{t}$, and repeatedly degrade the results back to an earlier timestep $t-1$, and then predicts $\Delta_{t-1}$, all the way to the final colorized images. In conclusion, Diffusing Colors presents an effective and novel diffusion-based image colorization framework, which overwrites the basic rules of iterative denoising processing in diffusion.

\section{Motivation of Hue-invariant FID}
\begin{figure}
    \centering
    \includegraphics[width=0.4\linewidth]{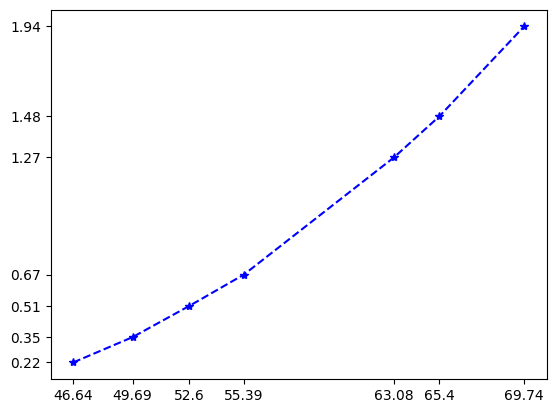}
    \caption{Colorfulness-FID trade-off under color enhancement on ground truth images. X-axis represents colorfulness metric and y-axis represents FID.}
    \label{col-fid}
\end{figure}

\begin{figure}
    \centering
    \includegraphics[width=0.8\linewidth]{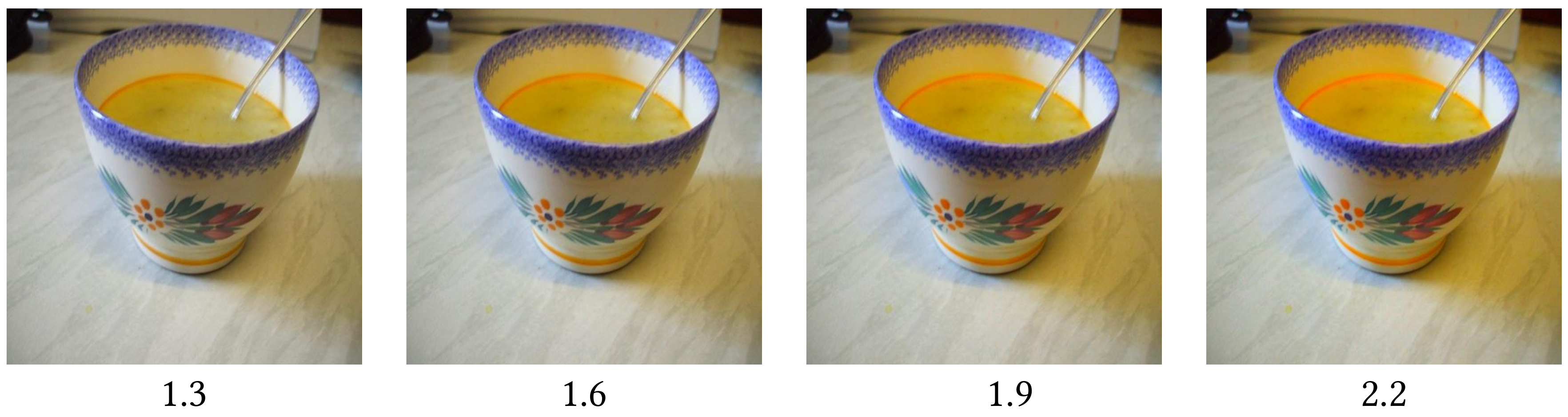}
    \caption{Visualization of YUV color enhancement under different scale factors, denoted by the captions of each subfigure.}
    \label{yuv_enhance}
\end{figure}

To clarify the impact of different color saturation on FID, we use \texttt{PIL.ImageEnhance} to linearly enhance color saturation of ground truth images in YUV color space under different scale factor, and calculate corresponding FID between color-enhanced images and original images, resulting in multiple pairs of (colorfulness, FID) data, as shown in Fig.~\ref{col-fid}, where x-axis represents colorfulness metric and y-axis represents FID. We further show some examples in Fig.~\ref{yuv_enhance} with different scale factor.

Notably, even natural ground truth images under different saturation lead to discrepancy on FID. To mitigate such impact of color saturation, we proposed Hue-invariant FID, or HI-FID, which further precisely measures local color distortions such as color overflows or color artifacts.

\section{Implementation Details}
\subsection{Training Configurations}
We set LoRA rank of UNet as 32 and the one of VAE as 4.
We train the whole model on a single NVIDIA RTX 4090 GPU in 4 epochs with batch size as 1 due to the limitation of GPU memory.
We believe increasing the batch size will further enhance Color-Turbo's performance, indicating our method has board potential improvements.
We use AdamW optimizer and set $\beta_1=0.9$ and $\beta_2=0.999$. The learning rate is set to 1e-5. 
To save memory cost, we utilize Accelerate~\cite{accelerate} and DeepSpeed~\cite{deepspeed} training architecture under float16 precision.

For the training of ControlNet-color, we conduct the whole ControlNet pipeline on the basis of Stable Diffusion v1.5.
We train the model on a single NVIDIA RTX 4090 GPU in 80k iterations with batch size as 64.
To save memory cost, we utilize Accelerate with gradient accumulation and checkpoints. We also use AdamW optimizer and set $\beta_1=0.9$ and $\beta_2=0.999$, setting the learning rate to 1e-5. We set the gradient accumulation step as 8 to balance the GPU memory and training time.

Both Color-Turbo and ControlNet-color use the training dataset adopted by~\cite{coco-lc}, which contains 125k images and corresponding text prompts totally.
\subsection{Calculation of Colorfulness Metric}
The colorfulness metric~\cite{hasler2003measuring} (CF) was proposed to evaluate the vividness of images, and has been widely adopted in image colorization field to measure the visual quality of generated color images, which can be formulated by
\begin{equation}
    CF = 0.3 \cdot \sqrt{\mu_{rg}^2 + \mu_{yb}^2} + \sqrt{\sigma_{rg}^2+\sigma_{yb}^2},
\end{equation}
where 
\begin{equation}
    \begin{aligned}
    \mu_{rg}=\mu(R-G), \mu_{yb}=\mu(0.5\cdot (R+G)-B), \\
    \sigma_{rg}=\sigma(R-G), \sigma_{yb}=\sigma(0.5\cdot (R+G)-B),
    \end{aligned}
\end{equation}
and $\mu(\cdot), \sigma(\cdot)$ represent the mean and standard deviation values respectively, $R,G,B$ denote RGB channels. 

\subsection{Calculation of FID and CLIP score}
We adopt \texttt{pytorch-fid}~\cite{pytorch-fid} to calculate FID metric. As for CLIP score, we adopt openai CLIP version\footnote{https://github.com/openai/CLIP} to extract embedding of images and texts.

\subsection{Filter protocal of BLIP captioner}
As mentioned at the main text, we leverage BLIP~\cite{blip} as an image captioner to produce essential text prompts as input. However, we experimentally find that BLIP sometimes produce confusing words with grayscale images, such as ``arafed", ``arafing", \textit{etc}. Therefore we simply filter out all such words with ``ara" prefix. Besides, we filter out some descriptions like ``black and white photo(s)". Fig.~\ref{blip_cleaner} shows Python code of BLIP's text cleaning.
\begin{figure}
    \centering
    \includegraphics[width=0.5\linewidth]{figures/blip cleaner.png}
    \caption{Python code of BLIP's text cleaning}
    \label{blip_cleaner}
\end{figure}

\section{Quantitative results of Inference time comparison}
\begin{table}[h]
\centering
\caption{Inference time comparison of diffusion-based methods.}
\label{Table:speed}
\resizebox{0.5\linewidth}{!}{
\begin{tabular}{c|c|c}
\toprule
Method & Resolution & Time (seconds per 100 images) \\ \midrule
L-CAD & 256$\times$256 & 2.53 \\
COCO-LC & 512$\times$512 & 6.42 \\
ControlNet-color & 512$\times$512 & 4.77 \\
Color-Turbo (Ours) & 512$\times$512 &\textbf{0.33} \\
\bottomrule
\end{tabular}
}
\end{table}

\section{More Quantitative Results with PSNR, SSIM, LPIPS Measurements}
We show PSNR~\cite{psnr}, SSIM~\cite{ssim} and LPIPS~\cite{lpips} results in Table~\ref{quan-psnr_uncond}, \ref{quan-psnr_cond}.
However, due to the inherent color ambiguity, we argue that paired data-based metrics such as PSNR, SSIM, LPIPS are not suitable to serve as quantitative measurements on colorization task. We summarize three-fold catastrophic drawbacks of those metrics as follows. 
\begin{itemize}
    \item \textbf{Unstable quantitative measurement.} Since these three metrics are sensitive to subtle disturbances in an image which have slight impact on overall color quality, the trend of these metrics on different dataset are strangely different. For example, InstColor~\cite{instColor} performs better on Extended COCO-Stuff dataset with higher SSIM and LPIPS than other evaluation datasets, but presents obviously lower PSNR. Likewise, L-CoIns~\cite{lcoins} on automatic task performs good PSNR on Multi-instance Dataset, but rather lower SSIM. Such inconsistency between different metrics will confuse researchers and hamper their practical usability.
    \item \textbf{Mismatch with users' preference.} By comparison with user study in the main text, CIC holds only $\sim$1\% user support, but achieve best PSNR, SSIM, LIPIS 5/9 on three different evaluation dataset. 
    \item \textbf{Unfair for better results.} Fueled with powerful diffusion models, some methods can even produce more vivid and plausible colorized images than ground truth images as shown at the main text and the following more visualized results. However, pair data-based metrics eliminates such advantages, as COCO-LC and ControlNet-color cannot achieve superior performance under such measurement.
\end{itemize}
\begin{table*}[h]
\caption{PSNR, SSIM and LPIPS results on automatic colorization task.}
\label{quan-psnr_uncond}
\resizebox{\linewidth}{!}{
\begin{tabular}{c|c|ccc|ccc|ccc}
\toprule
\multirow{2}{*}{Methods} & \multirow{2}{*}{category} & \multicolumn{3}{c|}{Extended COCO-Stuff   Dataset} & \multicolumn{3}{c|}{Multi-instances Dataset} & \multicolumn{3}{c}{ImageNet-val5k} \\
 &  & PSNR$\uparrow$ & SSIM$\uparrow$ & LPIPS$\downarrow$ & PSNR$\uparrow$ & SSIM$\uparrow$ & LPIPS$\downarrow$ & PSNR$\uparrow$ & SSIM$\uparrow$ & LPIPS$\downarrow$  \\ \midrule
CIC~\cite{cic} & \multirow{3}{*}{CNN-based} & 23.49 & \underline{0.9316} & \textbf{0.1689} & 23.86 & \textbf{0.9205} & \textbf{0.1512} & \textbf{24.24} & \underline{0.919} & \textbf{0.1532} \\
 InstColor~\cite{instColor} &  & 23.19 & 0.9279 & \underline{0.1716} & \textbf{24.07}	& 0.8855 & 0.1733 & 22.77 & 0.8812 & 0.1982 \\
 DISCO~\cite{DISCO} &  & 20.58 & 0.894 & 0.1952 & 20.84 & 0.8858 & 0.1995 & 20.92 & 0.8791 & 0.1802 \\ \cmidrule{1-11} 
 ColTran~\cite{coltran} & \multirow{4}{*}{Transformer-based} & 21.89 & 0.8949	& 0.2241 & 22.03 & 0.8961	& 0.2203 & 23.03 & 0.9062 & 0.202 \\
 CT2~\cite{ct2} &  & 22.56 & 0.9168	& 0.2103 & 22.66 & 0.9036 & 0.1884 & 23.16 & 0.9059 & 0.1875  \\
 ColorFormer~\cite{colorformer} &  & 22.87 & 0.906 & 0.1847 & 22.89 & 0.8961 & 0.192 & 23.08 & 0.8879 & 0.1594 \\
 DDColor~\cite{ddcolor} &  & 22.91 & 0.9119 & 0.1773 & 22.68	& 0.8866 & 0.1903 & 22.9 & 0.8949 & \underline{0.1644}  \\ \cmidrule{1-11}
 Deoldify~\cite{DeOldify} & \multirow{4}{*}{GAN-based} & 23.2 & 0.9254 & 0.1763 & 23.43 & 0.8946 & 0.2339 & 23.87 & 0.895 & 0.2354  \\
 HistoryNet~\cite{HistoryNet} &  & 21.76 & 0.9115 & 0.2034 & 22.39 & 0.9088 & 0.2238 & 22.59 & 0.9065 & 0.2318 \\
 ToVivid~\cite{tovivid} &  & 21.83 & 0.9027 & 0.2074 & 22.18	& 0.8928 & 0.2117& 21.98 & 0.8567 & 0.2115  \\
 ChormaGAN~\cite{ChromaGAN} &  & 22.7 & 0.9133 & 0.2029 & 22.94 & 0.9002 & 0.1829 & 23.29 & 0.9005 & 0.2124  \\
 BigColor~\cite{bigcolor} &  & 19.29	& 0.8635 & 0.2564 & 18.92 & 0.8236 & 0.2578 & 19.12 & 0.8253 & 0.2629  \\ \midrule
 L-CoDe~\cite{lcode} & \multirow{3}{*}{\thead{from scratch\\ cross-modality}} & 23.64 & 0.9251 & 0.1764 & 23.73 & 0.916 & 0.183 & 23.32 & 0.9113 & 0.2035  \\
 L-CoDer~\cite{lcoder} & & 23.28 & 0.9266 & 0.1802 & \underline{23.58} & 0.9133 & 0.1877 & 23.96 & 0.9137 & 0.1914  \\
 L-CoIns~\cite{lcoins} & & 23.02 & 0.9228 & 0.1864 & 24.02 & 0.9144 & 0.1797 & 23.56 & 0.9102 & 0.2001  \\ \cmidrule{1-11} 
 UniColor~\cite{unicolor} & \multirow{6}{*}{\thead{pretrained \\ cross-modality}} & 22.13 & 0.9101 & 0.1942 & 22.55 & 0.8987 & \underline{0.1701} & 21.76 & 0.8838 & 0.2235  \\
 L-CAD~\cite{lcad} & & \textbf{26.35} & \textbf{0.9373} & 0.1727 & 22.91 & 0.9038 & 0.1862 & 22.86 & 0.9011 & 0.202  \\
 COCO-LC~\cite{coco-lc} & & 23.97 & 0.9129 & 0.1963 & 22.76 & 0.9021 & 0.1971 & 22.63 & 0.9018 & 0.1964  \\
 ControlNet-color &  & 22.21 & 0.9055 & 0.2083 & 22.32 & 0.9141 & 0.2003 & 22.54 & 0.9103 & 0.2192  \\
 Color-Turbo ($\alpha=0.5$) & & 23.89 & 0.9043 & 0.2023 & 21.92 & 0.9042 & 0.213 & 22.92 & 0.9128 & 0.2069  \\
 Color-Turbo ($\alpha=1$) & & \underline{24.36} & 0.9076 & 0.1945 & 22.37 & \underline{0.9171} & 0.1998 & \underline{23.58} & \textbf{0.9201} & 0.1871  \\

\bottomrule
\end{tabular}
}
\end{table*}

\begin{table*}[h]
\caption{PSNR, SSIM and LPIPS results on language-based colorization task.}
\label{quan-psnr_cond}
\resizebox{\linewidth}{!}{
\begin{tabular}{c|c|ccc|ccc|ccc}
\toprule
\multirow{2}{*}{Methods} & \multirow{2}{*}{category} & \multicolumn{3}{c|}{Extended COCO-Stuff   Dataset} & \multicolumn{3}{c|}{Multi-instances Dataset} & \multicolumn{3}{c}{ImageNet-val5k} \\
 &  & PSNR$\uparrow$ & SSIM$\uparrow$ & LPIPS$\downarrow$ & PSNR$\uparrow$ & SSIM$\uparrow$ & LPIPS$\downarrow$ & PSNR$\uparrow$ & SSIM$\uparrow$ & LPIPS$\downarrow$  \\ \midrule
 L-CoDe~\cite{lcode} & \multirow{3}{*}{\thead{from scratch\\ cross-modality}} & 23.38 & 0.9255 & 0.1804 & 21.85 & 0.8132 & 0.2769 & 23.32 & \underline{0.9113} & 0.2035 \\
 L-CoDer~\cite{lcoder} &  & 24.08 & 0.926 & 0.1712 & 23.38 & 0.9138 & 0.1905 & 24.02 & \textbf{0.9144} & \textbf{0.1797} \\
 L-CoIns~\cite{lcoins} &  & 23.52 & 0.93 & 0.1726 & 22.34 & 0.8158 & 0.2643 & 23.56 & 0.9102 & 0.2001  \\ \cmidrule{1-11} 
 UniColor~\cite{unicolor} & \multirow{6}{*}{\thead{pretrained \\ cross-modality}} & 21.21 & 0.8955 & 0.2127 & 22.73 & 0.9023 & 0.1994 & 22.54 & 0.8913 & 0.2097  \\
 L-CAD~\cite{lcad} &  &\textbf{27.41} & \textbf{0.9408} &\textbf{0.1101} & 22.48 & 0.8973 & 0.1925 & 23.06	& 0.8991 & 0.1963  \\
 COCO-LC~\cite{coco-lc} &  & 24.26 & 0.9264 & 0.170 & 23.57 & \underline{0.9203} & 0.1932 & 24.21 & 0.911 & 0.1987  \\
 ControlNet-color &  & 23.72 & 0.918 & 0.1927 & 22.79 & 0.9172 & 0.2249 & 23.03 & 0.8892 & 0.2128  \\
 Color-Turbo ($\alpha=0.5$) &  & 23.92 & 0.9301 & 0.1676 & \underline{23.68} & 0.9193 & \underline{0.1702} & \underline{25.29} & 0.8928 & 0.1877  \\
 Color-Turbo ($\alpha=1$) &  & \underline{24.36} & \underline{0.9378} & \underline{0.1355} & \textbf{24.79} & \textbf{0.9362} & \textbf{0.1453} & \textbf{25.97} & 0.9012 & \underline{0.1798}  \\

\bottomrule
\end{tabular}
}
\end{table*}


\section{More Qualitative Results}
\begin{figure}[h]
    \centering
    \includegraphics[width=\linewidth]{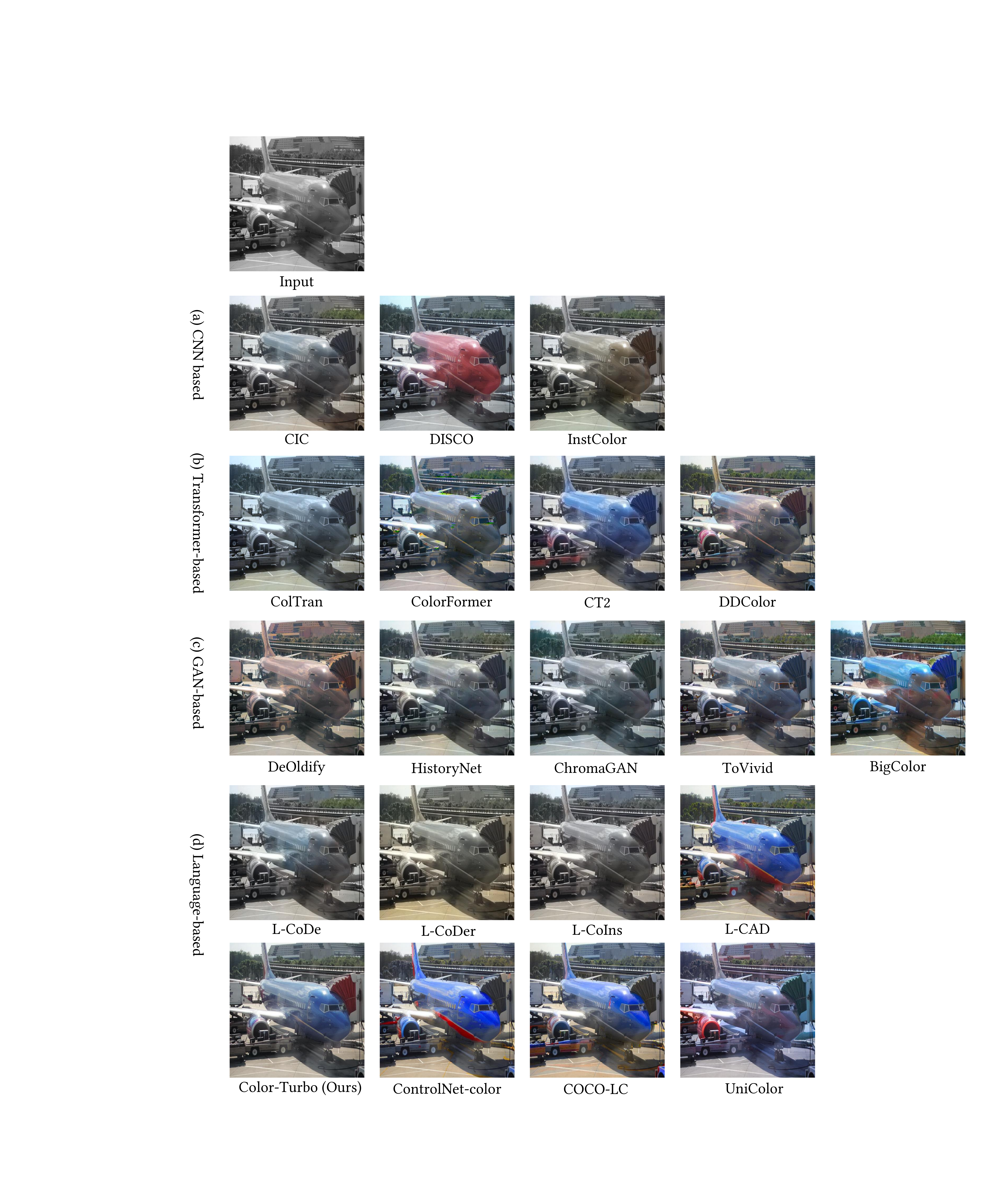}
    \caption{More visualized results on automatic colorization task}
    \label{supp_figures_uncond_1}
\end{figure}

\begin{figure}[h]
    \centering
    \includegraphics[width=\linewidth]{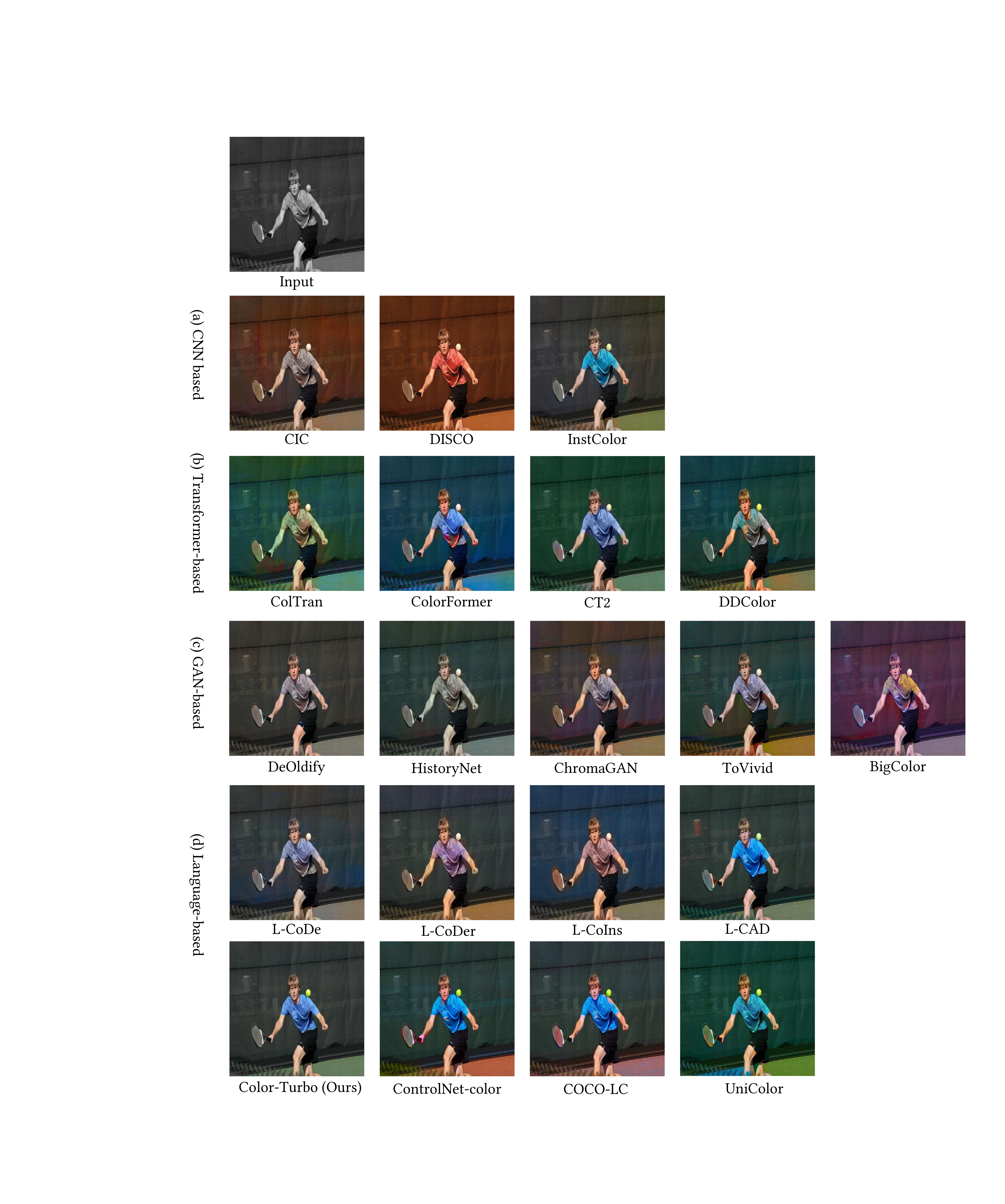}
    \caption{More visualized results on automatic colorization task}
    \label{supp_figures_uncond_2}
\end{figure}

\begin{figure}[h]
    \centering
    \includegraphics[width=\linewidth]{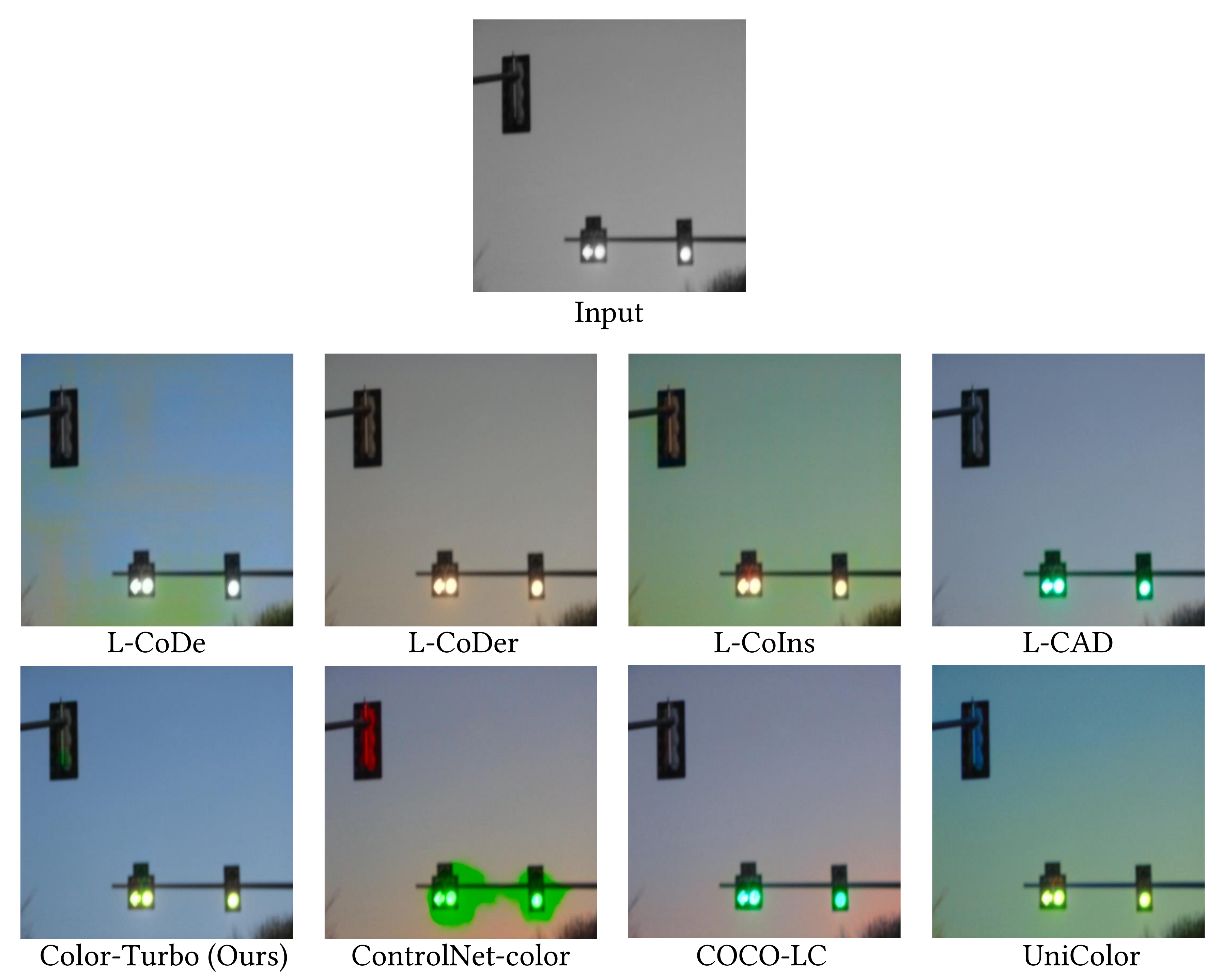}
    \caption{More visualized results on language-based colorization task. Prompt: ``a green traffic light suspended above a street".}
    \label{supp_figures_cond_1}
\end{figure}

\begin{figure}[h]
    \centering
    \includegraphics[width=\linewidth]{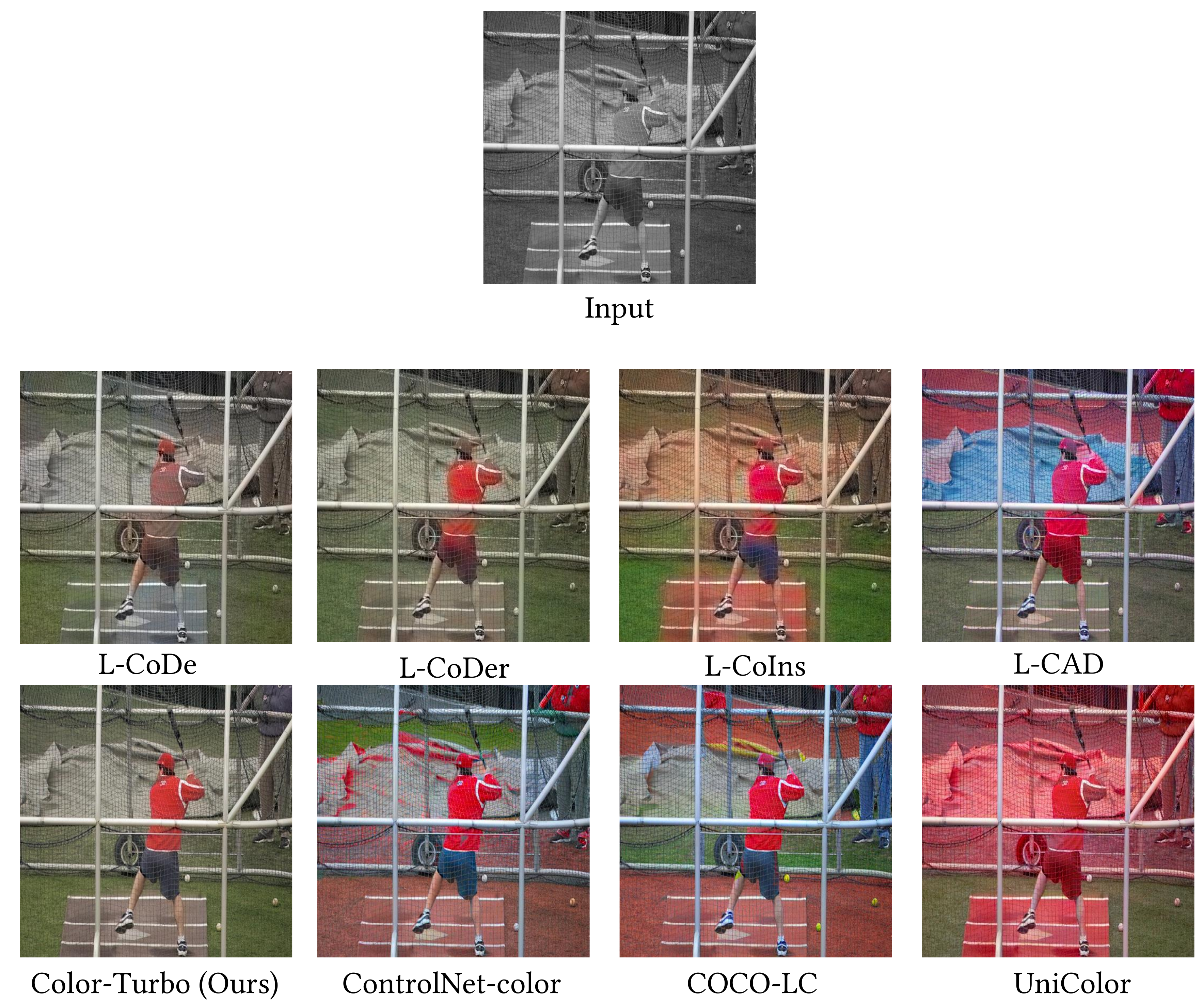}
    \caption{More visualized results on language-based colorization task. Prompt: ``a man in red jacket and shorts in a batting cage".}
    \label{supp_figures_cond_2}
\end{figure}

We shown more visualized results in Fig.~\ref{supp_figures_uncond_1}, \ref{supp_figures_uncond_2} on automatic colorization task and Fig.~\ref{supp_figures_cond_1}, \ref{supp_figures_cond_2}, \ref{supp_figures_cond_3}, \ref{supp_figures_cond_4} on language-based colorization task.

\newpage

\newpage

\newpage
\begin{figure}[h]
    \centering
    \includegraphics[width=\linewidth]{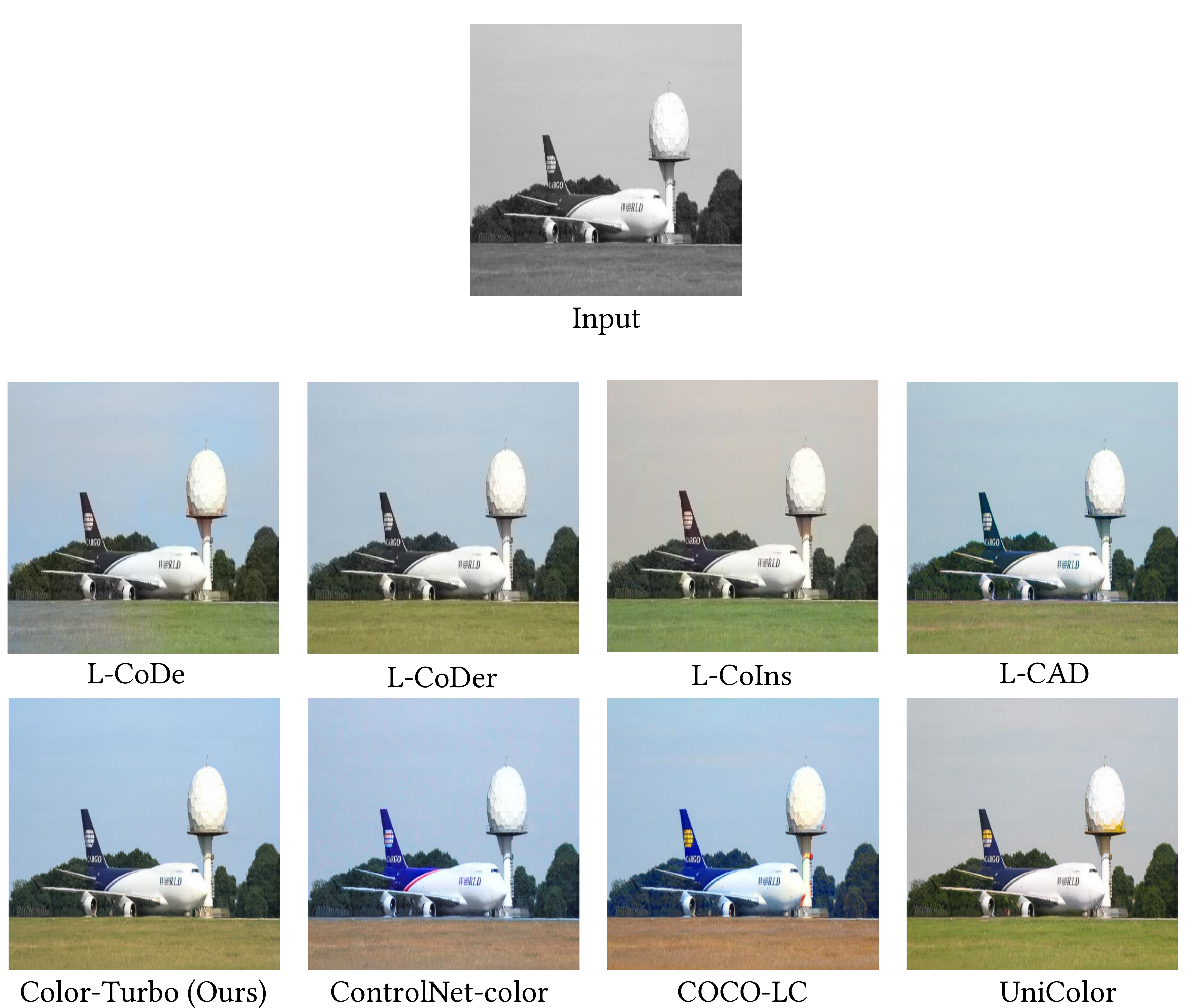}
    \caption{More visualized results on language-based colorization task. Prompt: ``a jet plane and a large white sphere".}
    \label{supp_figures_cond_4}
\end{figure}

\newpage
\begin{figure}[h]
    \centering
    \includegraphics[width=\linewidth]{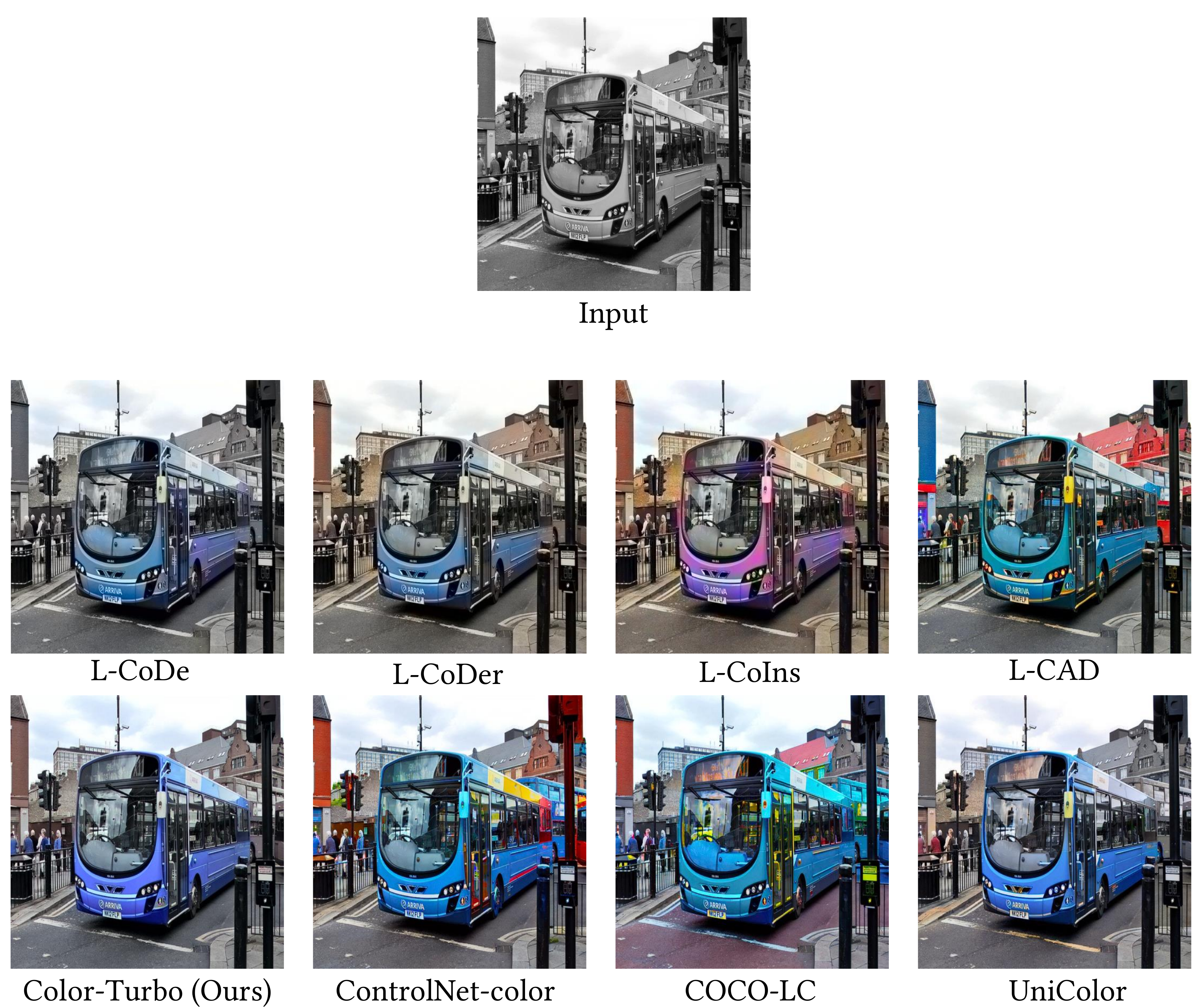}
    \caption{More visualized results on language-based colorization task. Prompt: ``a blue bus driving down a city street".}
    \label{supp_figures_cond_4}
\end{figure}

\newpage
\bibliographystyle{IEEEtran}
\bibliography{ms}

\ifCLASSOPTIONcaptionsoff
  \newpage
\fi